\documentclass[3p,fleqn,10p,authoryear]{elsarticle}

\usepackage[english]{babel}
\usepackage[utf8]{inputenc}
\usepackage[T1]{fontenc}

\usepackage{amsthm,amssymb}
\usepackage[fleqn]{mathtools}
\usepackage{booktabs,multirow,array}
\usepackage{graphicx}
\usepackage{tabu}
\usepackage{booktabs,caption,subcaption}
\usepackage{hyperref,setspace}
\usepackage[ruled,vlined]{algorithm2e}
\usepackage[nameinlink,capitalize]{cleveref}
\usepackage{lineno}

\graphicspath{{figures/}}

\hypersetup{
    colorlinks = true
}

\newdefinition{exm}{Example}
\newdefinition{lem}{Lemma}

\begin{document}
\begin{frontmatter}
    \title{A bi-partite generative model framework for analyzing and simulating large scale multiple discrete-continuous travel behaviour data}
    \author[1]{Melvin Wong}\corref{cor1}
    \ead{melvin.wong@ryerson.ca}
    
    \author[1]{Bilal Farooq}
    \ead{bilal.farooq@ryerson.ca}
    
    \address[1]{Laboratory of Innovations in Transportation, Civil Engineering Department, Ryerson University, 350 Victoria Street, Toronto, Ontario M5B 2K3, Canada}
    \cortext[cor1]{Corresponding Author.}

    \begin{abstract}
        The emergence of data-driven demand analysis has led to the increased use of generative modelling to learn the probabilistic dependencies between random variables.
        Although their apparent use has mostly been limited to image recognition and classification in recent years, generative machine learning algorithms can be a powerful tool for travel behaviour research by replicating travel behaviour by the underlying properties of data structures.
        In this paper, we examine the use of generative machine learning approach for analyzing multiple discrete-continuous (MDC) travel behaviour data.
        We provide a plausible perspective of how we can exploit the use of machine learning techniques to interpret the underlying heterogeneities in the data.
        We show that generative models are conceptually similar to the choice selection behaviour process through information entropy and variational Bayesian inference.
        Without loss of generality, we consider a restricted Boltzmann machine (RBM) based algorithm with multiple discrete-continuous layers, formulated as a variational Bayesian inference optimization problem.
        We systematically describe the proposed machine learning algorithm and develop a process of analyzing travel behaviour data from a generative learning perspective.
        We show parameter stability from model analysis and simulation tests on an open dataset with multiple discrete-continuous dimensions from a data size of 293,330 observations. 
        For interpretability, we derive the conditional probabilities, elasticities and perform statistical analysis on the latent variables. 
        We show that our model can generate statistically similar data distributions for travel forecasting and prediction and performs better than purely discriminative methods in validation.
        Our results indicate that latent constructs in generative models can accurately represent the joint distribution consistently on MDC data.
    \end{abstract}
    
\end{frontmatter}

% \linenumbers
\section{Introduction}
Large scale ubiquitous multidimensional travel data sources such as smartcard data or on-demand ride-sharing services provide enormous potential for travel behaviour analysts to implement new and innovative methods and algorithms for travel behaviour pattern forecasting \citep{ma2013mining,zheng2016big}.
In addition to size, these abstract data are also increasing in complexity, which necessitates data pruning or sub-sampling techniques to extract useful information and to improve estimation time at the cost of model accuracy.
Until recently, the most popular approach for travel behaviour modelling applications were hypothesis-driven discrete choice models (DCM). 
At the core, DCMs consist of defining a set of rules for Random Utility Maximization (RUM) \citep{train2009discrete}.
For instance, RUM have been been used in estimating route choice models with traffic network and socio-demographic information, including regret minimization \citep{chorus2008random}, prospect theory \citep{tversky2016advances} and the rational inattention model \citep{fosgerau2017discrete}.
Generative modelling proposes an alternative approach to analyzing travel behaviour data by constructing a model of the underlying distribution using unsupervised learning to generate new data with similar stochastic variations as the population \citep{stanislav2019generate}.
In contrast, DCM is optimized from maximum utility by estimating conditional probability distributions through a hypothesis-driven process with assumptions on the prior distributions.
When applied to travel behaviour datasets, the generative model behaves as an information processing constraint of the individuals as part of their decision process.
In this paper, we present our methodology using a machine learning algorithm that provides a systematic econometric analysis evaluation method, the possibility of identifying sources of heterogeneity associated with latent variables and model identifiability, essential elements in the transport analysis field.

% motivation
The benefits of using generative modelling are tied to behaviour theory and information processing cost in macroeconomic problems – generative models provide a more plausible framework for understanding selective and dynamic responses \citep{friston2007free}.
Previous work has provided a theoretical explanation to these interactions using artificial neural networks and how sensory information is reconstructed through generative modelling \citep{friston2007free,schwartenbeck2015evidence}.
The goal of this study is to present generative models as a behaviourally intuitive representation of travel decision making with an endogenous learning process. 
We argue that the main advantage of generative machine learning is that we can rely less on hypothesis-driven behaviour assumptions and representing decision perturbations beyond unobserved utility terms \citep{keane1997current}.
Recent developments in artificial neural network and learning algorithms have made it possible to estimate complex and non-rational behaviour (relaxation of IID assumptions) models that generalize better to various decision-making strategies \citep{matejka2015rational}. 
This paper offers a plausible perspective of how we can exploit the use of emerging machine learning techniques to model the behavioural processes before decision making actions.

We propose an extension for generative machine learning to specifically model multiple discrete-continuous (MDC) large-scale travel behaviour data.
We show that our proposed model can generate reasonably accurate data reconstructions, given suitable data observations and capacity for training.
Our proposed generative model provides a simple and intuitive mechanism for understanding the trade-offs between entropy and utility-maximizing behaviour by resolving uncertainty using variational Bayesian inference methods.

The main contributions of this paper are summarized as follows:
\begin{enumerate}
    \item We propose a bi-partite generative model to handle large travel behaviour datasets with MDC data types using an RBM learning algorithm;
    \item Systematically describe the machine learning framework used to train the generative model using a variational Bayesian inference objective function;
    \item Show how an information-theoretic model leads to economic behaviour compatibility that can be understood as: (a) lower evidence bound that depends on a variational free energy function, and (b) a measure of risk minimization that approximates the posterior distribution;
    \item Develop analytical methods to generate conditional probabilities, elasticities and latent variable distributions that can be used for interpretation and economic analysis.
\end{enumerate}

With the emergence of data-driven demand and services that use abstract forms of data, for example, social media data, there is a need to understand the underlying properties and correlation between ‘Big Data’ sources and choice actions to model travel behaviour using the potential of modern generative and deep learning techniques.
This paper aims to bridge the gap between traditional means of travel behaviour analysis dependent on identifiable variables and using abstract data which require machine learning techniques to extract useful information.
The novel approach tackles the problem of representing information heterogeneity in data-driven behaviour models using a joint distribution of discrete and continuous data.

This paper is organized as follows:
In section 2, we explain the background of the generative model and the variational Bayesian inference method.
In section 3, we describe our adaptations of generative machine learning methods, implementation on discrete and continuous travel behaviour datasets and optimization using variational Bayesian inference.
In section 4, we present the case study and results on large scale travel data.
Finally, discussions and conclusion are in section 5.

\section{Literature review}
Conventional DCM is used to estimate travel behaviour models from large scale multidimensional geospatial datasets, e.g. GPS systems \citep{menghini2010route,sobhani2018metropolis}.
However, missing or noisy data could lead to inaccuracy in model estimation and may require incorporation of latent variables.
In transportation, obtaining useful information from these datasets may be difficult because important trip details (mode choice, pricing, number of passengers, etc.) cannot be recorded directly from GPS data points \citep{shen2014review}.
Another obstacle is defining a generalized framework for incorporating latent variables or missing data points into multidimensional choice models.
Latent variables are essential in travel behaviour modelling as they capture behavioural perceptions related to uncertainty and describes the underlying mechanism of the choice selection process \citep{ashok2002extending}.
However, model specifications with complex distributions may not produce an identifiable closed-form solution for maximum likelihood estimation.
For the above reasons, researchers have implemented Monte Carlo methods and variational Bayesian inference for analytical approximations to incorporate mixed distributions and choice dynamics into the model estimation process \citep{bhat2011maximum,vij2015big}.

Variational Bayesian inference combines prior knowledge and empirical evidence to resolve uncertainty and adapt to noisy datasets through data-driven algorithms such as neural networks and generative models \citep{ghahramani2015probabilistic}.
Variational Bayesian inference methods are widely used in machine learning with successful applications in data mining and sentiment analysis \citep{witten2011data,wang2017generative}.
In classical Bayesian modelling, the posterior distributions are usually estimated by simulation or sampling-based methods.
A commonly employed sampling-based algorithm for travel behaviour dataset is the Markov Chain Monte Carlo (MCMC) algorithm where the posterior distribution is simulated by drawing repeated samples from a Markov Chain until convergence \citep{hastings1970monte}.
The stationary distribution of the Markov chain represents the posterior distribution.

In recent years, MCMC algorithm has played an important role in travel behaviour modelling problems in transportation, with successful applications in agent-based simulations \citep{lee2009new}, hybrid choice models \citep{allenby1998marketing,daziano2013incorporating}, and population synthesis \citep{farooq2013simulation,sun2015bayesian,saadi2016hidden}.
However, in order to match the asymptotic efficiency of maximum likelihood, MCMC draws must grow at a rate faster than the square root of the number of agents \citep{train2009discrete,bhatnagar2011computational}.
With complex mixing distributions, convergence may not be guaranteed in a reasonable time, resulting in poor estimation.
This makes sampling-based estimation methods infeasible beyond relatively simple models and small datasets for obtaining accurate results.
This problem has led to the development of convergence testing methods to assess model precision \citep{bhatnagar2011computational}.
Another viable approach is the iterative Expectation-Maximization (EM) algorithm for posterior estimation \citep{dempster1977maximum}.
Although EM algorithm may be useful in small datasets and for incomplete data, the rate of mixing is also known to be extremely slow in some cases \citep{bhat1997endogenous,train2008algorithms}.

\subsection{Conventional MDC model estimation approaches}
\label{subsec:conventional}
The conventional hypothesis-driven approach for MDC modelling is primarily by the multiple discrete-continuous extreme value (MDCEV) model \citep{bhat2008multiple}.
It incorporates a non-linear function in the utility structure to account for choice substitutions, continuous consumption and multiple alternatives.
In MDCEV model, multiple constraints are pre-defined, hypothesis-driven based utility function.
There is the assumption on MDCEV that a single baseline utility influences both discrete and continuous consumption.
Although this has been expanded recently by incorporating different utility functions for discrete and continuous options \citep{bhat2018flexible}.
Other models for estimating MDC include the \textit{translated quadratic non-linear additive model} which provides corner solutions and diminishing marginal utility.
This has been used in modelling consumer choices with multiple purchase variety \citep{kim2002modeling}.

Abundant sources of travel behaviour datasets are becoming available via new sources like social media, smartphone apps, and communication networks. 
There is a need for new approaches that are specifically designed for these large datasets. 
Our current work differs from hypothesis-driven approaches in which we develop a generative model with a joint distribution accounting for latent correlation effects in large datasets.
The result is a data-driven generative model described by the underlying latent behavioural distribution, and the solution entails finding the model parameters that best replicate the outcomes.
We develop the estimation procedure using a Gibbs sampling based gradient descent method, typically used in machine learning.

\subsection{Existing developments of generative modelling in transportation}
One of the critical issues in discrete choice model design is the assumption that observations are drawn independently, although this assumption of often always violated in real-world problems.
Alternatively, this problem can be handled by considering a more flexible model with a richer set of random variables with data-driven distributions that allows practitioners to describe a model that best represents the behaviour of the population.

In transport modelling, several studies have been conducted that investigate how probabilistic models can be effectively leveraged to model spatial-temporal data through Bayesian inference techniques.
Probabilistic models have been described to be a form of `transfer learning scheme' instead of traditional learning where calibration is done on a single source of labelled data \citep{anda2017transport}.
Transfer learning enables relaxation of various assumptions in the modelling process and being able to reconstruct new and unseen observations from the joint probability which is useful for exploiting and extracting non-survey based data, e.g. social media data, that has little direct correlation to travel behaviour.
In transport studies, model-based machine learning approaches such as generative modelling are primarily used for classification of unseen observation by identifying the latent variables that describe some contextual information not captured in the data \citep{sun2017discovering}.
Latent Dirichlet Allocation (LDA) \citep{blei2003latent} is another popular variation of generative modelling that is commonly used to analyze structure in the data without prior labels, for example, the discovery of activity patterns in trip modelling \citep{huynh2008discovery,hasan2014urban}.

Probabilistic Graphical Models (PGMs) describes the representation and structure of probability distributions compactly and intuitively by encoding the independence assumptions and causality between random variables in the factorized graph edges \citep{peled2019model}.
Each edge connection corresponds to the strength of direct dependence between the random variables, and each random variable can be constructed as a conditional model given the other variables and the corresponding edges.
PGMs have been used for traffic simulation by representing traffic links as the graph edges and estimating the model using a first-order spatial Markov model \citep{muralidharan2011probabilistic}.
\citep{wheeler2016factor} developed a PGM for realistic highway scenes by modelling vehicles as nodes and interactions between vehicles as factor graph edges.
By generating novel `path' probabilities between random variables, PGMs can model all types of interactions and correlations that can best represent the underlying properties of discrete and continuous data.

\subsection{Generative modelling using artificial neural networks}
\label{subsec:generative}
Generative models are used to learn a representation of a dataset as a joint distribution over the observed variables.
The joint distribution analyzes the extracted information without relating it to the observers' prior knowledge, and these subjective measures are based on so-called information criteria, e.g., Akaike's information criterion or Shannon entropy \citep{akaike1992information}.
Subjective measures consider additional knowledge about the observation such as novelty, counter-intuitive behaviour or familiarity.
Existing discrete choice models are based on such measures to represent latent behavioural information about the traveller's behaviour such as latent class (LC) models, Mixed logit (ML) and integrated choice and latent variable (ICLV) models \citep{shen2009latent}.

Early statistical methods used generative modelling for dimensional reduction such as principal component analysis (PCA), k-means clustering and linear discriminant analysis (LDA).
PCA can be used as a simple dimensional reduction tool that relies on linear assumptions where each dimension (PCA latent variable) is highly correlated to each other.
However, abstract data sources may not possess these properties and are more likely to be noisy, complex and have multiple non-linear correlations. 
In order to sufficiently capture non-linear variations in the data, deep learning techniques can be applied.

Recently, more powerful forms of generative models are based on neural networks and have been widely used in applications such as population synthesis, semantic analysis and recommendation systems.
Some of these generative models include restricted Boltzmann machines (RBM), generative adversarial nets (GAN) and variational autoencoders (VAE) \citep{goodfellow2016deep}.

RBMs are the earliest and most simple form of parametric generative models that perform representation learning by fitting the neural network model to the data.
RBMs are utilized as building blocks for constructing deep artificial neural nets such as Deep Belief Nets (DBN) \citep{salakhutdinov2015learning}.
Inference in RBM generative models is difficult; thus, efficient training algorithms were introduced to approximate the inference procedure \citep{hinton2002training}.
The general training process for RBM is a pairwise contrastive divergence algorithm which is bi-directional to allow up and downstream propagation of network weights.
Synthetic data can be sampled from the trained generative model that have similar statistical properties as the input dataset.
Compared with PCA or clustering based modelling approaches, RBMs have shown a strong capacity to model joint distributions and have been successfully applied to capture spatial-temporal patterns \citep{taylor2009factored}.
The RBM generative model restricts lateral connections within layers, which provides independent and identically distributed (IID) assumption about the observed and latent variables.
For prediction and forecasting, RBMs are typically used for learning latent features followed by either a generative simulation-based classifier or directly as a multi-layer neural network classifier \citep{wong2018discriminative}.

Other generative models such as VAEs are used to perform non-linear mapping of the input variables to `encodings' by compression and marginalizing out noisy data as part of the training process \citep{kingma2016improved}.
The `encodings' capture the most meaningful information of the data, similar to a clustering algorithm.
Estimation of VAE requires layer-wise training by optimizing the lower bound of a variational Bayesian inference objective function by applying a gradient-based updating rule.
GANs are another type of generative model that trains a generator and discriminator in the neural network simultaneously.
The discriminator attempts to distinguish between the real data and the generated data and minimizes the error of differentiating real from synthetic data.
This method is designed to be used for semi-supervised learning and is commonly implemented on computer vision and image classification tasks \citep{goodfellow2016deep}.

\subsection{Model optimization algorithms}
\label{subsec:model_opt}
The approach to solving the optimization problem in neural networks is to apply gradient descent via a backpropagation learning algorithm to calculate the gradients w.r.t. the likelihood function \citep{hinton2002training}.
This formula for gradient descent is applied to the variational inference algorithm in a generative model based on the principle of energy minimization \citep{schwehn2010using}.
A symmetric parameterized model such as the RBM uses a Gibbs sampler starting at some random data point that would allow the neural network to update the parameters until convergence is reached.
The procedure is known as \textit{blocked Gibbs sampling} by alternating updates between `visible' and `hidden' neurons.
However, the sampling approach requires running a Markov chain until convergence.
An approach using \textit{contrastive divergence} approximates the optimization problem by replacing the energy minimization gradient function by a fast approximate \citep{hinton2012practical}.

The objective of generative models is to learn meaningful ways to represent the input data through a subset of underlying latent variables.
This information processing architecture was suggested as a representation of behavioural stimuli \citep{friston2007free}.
It treats choice behaviour the same way as the rational inattention model, which depends on the context formed by prior beliefs \citep{matejka2015rational}.
Several studies have shown the superior performance of generative models in solving challenging decision-making problems over typical discrete choice and discriminative neural network.
To the best of our knowledge, the use of generative learning is limited to image and video data to capture motion and dynamics.
Here, we extend our previous work on RBM based single discrete choice and latent variable models \citep{wong2018discriminative} to incorporate multiple discrete-continuous choices.
We also propose a generic algorithm for estimating MDC models using generative machine learning.
The trained model is used to generate conditional samples and then used to perform classification tasks as well as travel behaviour prediction.

\section{Proposed generative machine learning approach}
\label{sec:ml}
In this section, we describe our adaptations of current machine learning methods, introduce our generative bi-partite framework for modelling MDC data and the associated model optimization algorithm.
A list of notations used throughout this paper is given in \cref{tab:notations}.

\begin{table}[!ht]
    \centering
    \begin{tabu}{X[0.3,c] X[0.7]}
        \toprule
        Notations &  Description \\
        \midrule
        $\mathbf{x}$ & set of input variables $x_1,x_2,...,x_K$ \\
        $\mathbf{s}$ & set of latent variables $s_1,s_2,...,s_J$ \\
        $\mathcal{H}[x]$ & entropy of $x$ \\
        $D_{KL}[a||b]$ & Kullback-Leibler divergence of $a$ from $b$ \\
        $\mathcal{F}$ & variational free energy \\
        $E(\mathbf{x})$ & energy of $\mathbf{x}$ \\
        $\langle x \rangle_q$ & expected value of $x$ over distribution $q$\\
        $\sigma(x)$ & sigmoid function operator $(1+e^{-x})^{-1}$ \\
        $\mathcal{N}(W,\Sigma^2)$ & Gaussian distribution with mean $W$ and variance $\Sigma^2$ \\
        $\nabla_\theta (f)$ & gradient of function $f$ w.r.t. $\theta$ \\
        $\eta$ & stochastic gradient descent rate. Note: $\eta < 1$\\
        \bottomrule
    \end{tabu}
    \caption{Notations}
    \label{tab:notations}
\end{table}

\subsection{Generative bi-partite model}
Conventional DCM methods often face difficulties in estimating large datasets with MDC choice outputs due to exponentially increasing choice set selection \citep{bekhor2006evaluation}.
Furthermore, the complexity of estimating DCM increases when incorporating hidden variables, requiring additional variational parameters while making model inference intractable and impractical.
One approach we can use is to approximate each unobserved component with a point estimate. 
However, we cannot quantify the uncertainty or confidence interval of these hidden variables.
The other approach is to find a joint distribution of the hidden and observed components and perform Bayesian analysis -- this usually results in an intractable integral.
The core function of generative machine learning solves the two problems by computing the integral through optimization of a variational free energy objective function and uses probabilistic Bayesian techniques to obtain the parameters of the model.

\begin{figure}
    \centering
    \includegraphics{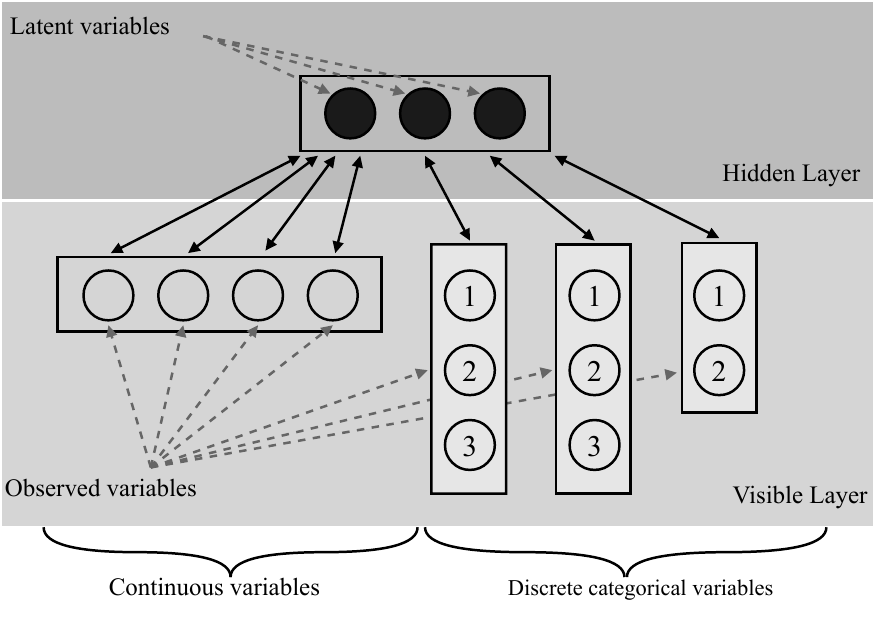}
    \caption{The generative bi-partite framework. The visible layer represents the input discrete and continuous data. The hidden layer represents the stochastic latent variables derived from the RBM learning algorithm. Bi-directional arrows indicate information passing in both directions. The hidden layer can be used to generate new data with similar statistical properties as the input.}
    \label{fig:rbm_schematic}
\end{figure}

Our proposed solution is a generative bi-partite graph framework that models the underlying processes that are likely to generate the data.
The assumption is that large amounts of data are available that can represent the true population behaviour.
See \cref{fig:rbm_schematic} for an illustration of the model.
First, we consider the joint distribution given as $p(\mathbf{x,s})$ over the set of \textit{binary} hidden random $\mathbf{s}=s_{1:J}\in\{0,1\}$ and observed $\mathbf{x}=x_{1:K}\in\mathbb{R}^{\mathcal{D}}$ variables.
We specify a prior distribution $p(\mathbf{s})$ about the hidden variables and quantify how $\mathbf{x}$ relates to $\mathbf{s}$ with the likelihood function $p(\mathbf{x|s})$.
Applying the Bayes' rule, we obtain the posterior distribution:

\begin{equation}
p(\mathbf{s|x}) = \frac{p(\mathbf{x,s})}{p(\mathbf{x})} \propto p(\mathbf{x|s})p(\mathbf{s})
\end{equation}

where $p(\mathbf{s})$ is the hidden layer distribution, e.g., Bernoulli, multinomial or normal, that are the latent priors, and conditional densities $p(\mathbf{x|s})$ are the likelihood components of the Bayesian model.
If the latent priors are tractable, the likelihood component may have $D_{\textrm{cont}}$ continuous and $D_{\textrm{cat}}$ discrete categorical components such that $\mathbf{x}$ can take the following dimensions:

\begin{equation}
\mathbf{x}_D = (\underbrace{x_1, ..., x_{D_{\textrm{cont}}}}_{\textrm{continuous}}, \underbrace{x_{D_{\textrm{cont}+1}}, ..., x_{D_{\textrm{cont}} + D_{\textrm{cat}}}}_{\textrm{discrete}})
\end{equation}

For categorical dimensions, we can apply a multinomial logistic distribution of $k$ possible alternatives represented by the vector $x_{D_{\textrm{cat}}} = (x_{D_{\textrm{cat}_1}},...,x_{D_{\textrm{cat}_k}})$ with $x_{D_{\textrm{cat}_k}}=1$ if the $k$ alternative for variable $x_{D_{\textrm{cat}}}$ is chosen.
The multinomial distribution is defined by:

\begin{equation}
p(x_{D_{\textrm{cat}_k}}=1) = \frac{e^{f_k(\mathbf{s}; \theta)}}{\sum_{k'} e^{f_{k'}(\mathbf{s}; \theta)}}
\end{equation}

The continuous multivariate component of this vector can be modelled with a normal distribution where $x_{D_{\textrm{cont}}}$ is drawn from a Gaussian $\mathcal{N}(W, \Sigma^2)$.
If $x_{D_{\textrm{cont}}}$ is not lower bound, the resulting function may generate negative values.
To distinguish between positive only values in travel behaviour data, e.g. speed, distance, a \textit{stepped sigmoidal} function can be used for generating positive real valued data:

\begin{equation}
\sum_{i=1}^{\infty} \sigma(\mathbf{s}-i) \approx \ln(1+e^{s})
\end{equation}

The sum of $\sigma(\mathbf{s}-i)$ components represents an infinite set of binary logistic models with shared weights and fixed constant offsets.
Applying this formulation increases the capacity of the logistic model to express a broader range of positive linear values but retains the same closed-form derivative and the same number of parameters.
It can also be further approximated with the function $\ln(1+e^{s})$.
This method has been used in the past to develop models such as the Infinite RBM and Rate-coded RBM in generative machine learning \citep{teh2000rate,cote2016infinite}.

As the hidden layer represents a fully distributed mixture model, the model can be considered a mixture model with $2^J$ components with $K+J+KJ$ parameters.
This representation of travel behaviour data makes it attractive because the complex correlations between observed variables and events as a result of interaction can be captured by a one or combination of multiple latent variables in the least number of additional parameters, as opposed to conventional Mixed Logit or latent class model.
We refer to \ref{appx:generation} for a detailed mathematical explanation on variable correlations among MDC choices and conditional probability generation.

\subsection{Variational Bayesian inference}
The marginal distribution of $\mathbf{x}$ can be obtained by integrating the joint distribution: $p(\mathbf{x})=\int_{\mathbf{s}} p(\mathbf{x|s})p(\mathbf{s}) d\mathbf{s}$.
We are interested in obtaining the posterior belief $p(\mathbf{s|x})$ that depends on the data to know how $p(\mathbf{x})$ are distributed.
Assuming that the data are conditional upon the hidden variables, the maximum likelihood of the data, i.e. $\operatorname*{arg\,max}_\theta \ln p(\mathbf{x})$ may be difficult as we require the integral to be tractable.
In most cases, it is difficult to compute in closed form and approximations are required.
A popular method of approximating the posterior is through the MCMC algorithms \citep{neal1993probabilistic}. However, such algorithms have a high computational cost and are more suited for well-structured small samples.
By starting from some arbitrary initial distribution $q(\mathbf{s}_0)$, a stochastic transitional distribution $\mathbf{s}_t \sim q(\mathbf{s}_t|\mathbf{s}_{t-1},\mathbf{x})$ is applied iteratively and the outcome $\mathbf{s}_T$ converges asymptotically to the exact posterior $p(\mathbf{s}|\mathbf{x}) \approx q(\mathbf{s}_0|\mathbf{x}) \prod_{t=1}^T q(\mathbf{s}_t|\mathbf{s}_{t-1},\mathbf{x})$.
The downside of this is that with MCMC, we do not know how many iterations are sufficient and finding the right posterior approximation may be difficult with large datasets and complex distributions.

Alternatively, it has been shown that the contrastive divergence algorithm works well on large datasets that may not be well-structured (\cref{subsec:model_opt}).
Variational Inference methods provides a better alternative to such problems by optimizing a more straightforward function that approximates the posterior faster than conventional sampling methods. 
Recent work have suggested that factorizing a variational distribution into common probability distributions works in practice to approximate the exact posterior, for example, in Mixed Logit models \citep{krueger2019variational}.
It has also been shown that for random utility-based choice models, the variational error is negligible and variational inference shows asymptotic behaviour \citep{braun2010variational}.
First, we posit that there is a tractable distribution $q(\mathbf{s})$ that approximates the exact posterior $p(\mathbf{s|x})$.
To find $q(\mathbf{s})$, we search over the set of distributions that minimizes the Kullback-Leibler (KL) divergence objective function:

\begin{equation}
    \begin{aligned}
    & \operatorname*{arg\,min} & & D_{KL}[q(\mathbf{s})||p(\mathbf{s|x})] \\
    & \hspace{1em} s.t.        & & \frac{p(\mathbf{s|x})}{q(\mathbf{s})} > 0, \\
    &                          & & D_{KL}[q(\mathbf{s})||p(\mathbf{s|x})] = 0 \iff q(\mathbf{s}) = p(\mathbf{s|x})
    \end{aligned}
    \label{eq:obj-kld}
\end{equation}

where \(D_{KL}[q(\mathbf{s})||p(\mathbf{s|x})] = -\int_\mathbf{s} q(\mathbf{s})\ln \frac{p(\mathbf{s|x})}{q(\mathbf{s})} d\mathbf{s}\). 
If no assumptions are made, then the equation is minimized when $q(\mathbf{s})=p(\mathbf{s|x})$.
The key benefit for variational Bayesian inference is that we can choose a restricted class of density distributions (partitions) for $q(\mathbf{s})$ which are simple enough for computational efficiency but flexible enough to capture the posterior distribution.

A simplifying assumption of $q(\mathbf{s})$ is that each of the partitions is independent and we can find a formula that computes $q(s_1,s_2,...,s_J)$ using the values of the observed input data.
This assumption means that the probabilities form an intersection of densities, which is an efficient way of modelling high-dimensional data while satisfying low-dimensional constraints \citep{bishop2006pattern}.
In comparison to latent class models, this translates adding contributions in the log domain, rather than in the probability domain.
The model can accommodate for a `no option' edge case in the probability density where a component has zero contribution (negative infinite energy) \citep{hinton2012practical}.
We factorize $q(\mathbf{s})$ by taking the product over independent latent variable densities:

\begin{equation}
    q(\mathbf{s}) = \prod_{j=1}^J q(s_j) \approx \prod_{j=1}^J
                    p(s_j|\mathbf{x}),\hspace{1em}
                    \mathbf{s} = \{s_1,s_2,...,s_J\}
\end{equation}

Each latent variable density $p(s_j|\mathbf{x})$ is a product of expert (PoE)  model.
The PoE distribution produces a model with marginal independent hidden states by specifying independent expert priors \citep{hinton2002training}.
If we assume each expert is a tractable distribution with a closed-form solution (e.g., logit or exponential), the generative model can be computed efficiently.
However, the objective function in \cref{eq:obj-kld} requires the computation of the marginal likelihood function $p(\mathbf{x})$ and $\ln p(\mathbf{x})$.
By applying a change-of-measure technique to the objective function and using Bayesian inference, we obtain:

\begin{eqnarray}
D_{KL}[q(\mathbf{s})||p(\mathbf{s|x})] &=& \int q(\mathbf{s})\ln q(\mathbf{s}) d\mathbf{s} - \int q(\mathbf{s})\ln p(\mathbf{s|x})d\mathbf{s}\\
&=& \int q(\mathbf{s})\ln q(\mathbf{s}) d\mathbf{s} -\int q(\mathbf{s})\ln p(\mathbf{x,s}) d\mathbf{s} +  \ln p(\mathbf{x}) \int q(\mathbf{s})  d\mathbf{s}\\
&=& -\mathcal{F} + \ln p(\mathbf{x}) \label{eq:elbo_f}
\end{eqnarray}

\noindent where $\int q(\mathbf{s})  d\mathbf{s}=1$, the expectation $\langle f(x)\rangle_q=\int f(x)q(x) dx$ and $\mathcal{F}$ is the variational free energy and can be expressed as:

\begin{equation}
\mathcal{F} = \langle\ln p(\mathbf{x,s})\rangle_q-\langle\ln q(\mathbf{s})\rangle_q = \langle\ln p(\mathbf{x,s})\rangle_q + \mathcal{H}[q]
\end{equation}

In practice, the Gibbs sampling algorithm is used to optimize the solution of the variational free energy objective function \citep{carreira2005contrastive}.
The variational free energy lower bounds the partition function $\ln p(\mathbf{x}) \geq \mathcal{F}$ for any $q(\mathbf{s})$, since $D_{KL}\geq 0$ holds which can be derived through the Jensen's inequality \citep{mackay2003information}.
We also note that $-\langle\ln q(\mathbf{s})\rangle_q=\mathcal{H}[q]$ is the entropy of the approximating distribution $q$ and $\langle\ln p(\mathbf{x,s})\rangle_q$ is the expected energy of the joint distribution.
Therefore, maximizing the variational free energy is equivalent to minimizing the KL divergence: \(\operatorname*{arg\,min} D_{KL}[q(\mathbf{s})||p(\mathbf{s|x})] = \operatorname*{arg\,max} F\).

The variational free energy indicates that decision makers are compelled to maximize both expected utility $\langle\ln p(\mathbf{x,s})\rangle_q$ and information $\mathcal{H}[q]$.
In purely econometric (utilitarian) choice models, independence of irrelevant alternatives holds and a rational decision-maker would always choose the alternative with the highest utility.
However, it is generally known that irrational behaviour plays a significant role in choice selection \citep{matejka2015rational,fosgerau2017discrete}.
In this context, incorporating KL divergence as a generalized measure of uncertainty in the model accounts for the variance over the utilities of the choices.
This is also known in some literature as risk-seeking or risk-avoiding behaviour \citep{chorus2008random,schwartenbeck2015evidence}. 
Next, we develop the parameter estimation procedure for the proposed generative model.

\subsection{Learning algorithm}
Standard learning algorithms for generative models utilize a stochastic gradient descent method for optimizing the objective function.
Assume that an arbitrary Gibbs-Boltzmann energy function is given by $E(\mathbf{x,s};\theta)$ where $\theta$ represents the model parameters.
The energy in this context describes a value that is assigned to a state of the system.
The energy curve is continuous, and the state(s) with the lowest energy corresponds to the highest probability.
We relate the RBM energy function to utility, where the inverse of utility is the energy, but states have both independent observed and latent variables.
Then the generative model is a joint probability distribution over the observed and latent variables in a configuration given by the Boltzmann probability distribution:

\begin{equation}
    p(\mathbf{x,s})=\frac{e^{-E(\mathbf{x,s})}}{\sum_{x,s} e^{-E(\mathbf{x,s})}}
    \label{eq:joint_pdf}
\end{equation}

\begin{figure}
    \centering
    \includegraphics[width=0.4\textwidth]{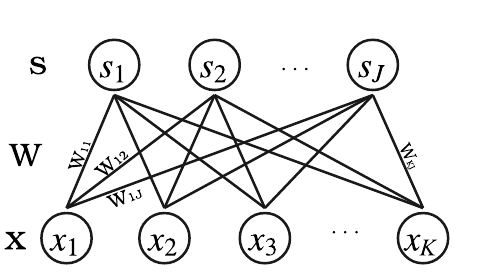}
    \caption{Graphical illustration of an RBM with connections represented by $\mathbf{W}$ between hidden $\mathbf{s}=(s_1,s_2,...,s_J)$ and visible layer $\mathbf{x}=(x_1,x_2,..,x_K)$. The connections are undirected, and the weights are the strength of the connections. Weight updates are performed bi-directionally in every batch step.}
    \label{fig:rbm}
\end{figure}

Illustrated in \cref{fig:rbm}, we express the RBM as a bi-partite graph of a visible and a hidden layer connected by a weight matrix.
These are considered as unsupervised learning methods, whereby there are no category labels or output values for model optimization.
RBM models are stochastic rather than deterministic: latent variables are randomly sampled according to a joint distribution specified by the model.
Let $\mathbf{W}\in\mathbb{R}^{K\times J}$ be the weight matrix connecting the hidden layer $\mathbf{s}=(s_1,s_2,...,s_J)$ and visible layer $\mathbf{x}=(x_1,x_2,..,x_K)$.
The magnitude of $\mathbf{W}$ measures the strength of the connection between two units.
The interaction between the two layers defines the energy function:

\begin{equation}
    E(\mathbf{x,s}) = - \mathbf{x^\top W s} - \mathbf{b^\top x} - \mathbf{c^\top s}
    \label{eq:expected_energy}
\end{equation}

The marginal of the visible layer is $p(\mathbf{x})=\sum_s p(\mathbf{x,s})$.
$\mathbf{b}$ and $\mathbf{c}$ are the parameters for the visible and hidden layer respectively towards the joint distribution density (\ref{appx:generation}).
The variational free energy objective is the lower bound approximation to the marginal log-likelihood.
From \ref{eq:elbo_f}, KL divergence is always positive, hence:

\begin{equation}
    \ln p(\mathbf{x}) \geq F
\end{equation}

The objective is to compute $q(\mathbf{s})$ that minimizes the KL divergence, which yields the variational density as an approximate to the posterior $q(\mathbf{s}) \approx p(\mathbf{s|x})$:

\begin{equation}
    \begin{aligned}
    &q(\mathbf{s}) := \max_{q(\mathbf{s})} F \iff \\ 
    &\nabla_{q(\mathbf{s};\theta)} F=0, \textrm{  for any } \theta^* \in \underset{x\in \mathbb{D}}{\operatorname*{arg\,max} } \ln p(\mathbf{x;\theta^*})
    \end{aligned}
\end{equation}

\noindent Using the definition of thermodynamic free energy in bounded rational decision making process \(F=U-T\mathcal{H}\), where \(U\) is the expected utility (energy), \(T\) is the temperature constant (\(T=1\)) and \(\mathcal{H}\) is the entropy \citep{collell2015brain,ortega2013thermodynamics}, we obtain the following derivative of the free energy:

\begin{eqnarray}
    \nabla_{q(\mathbf{s};\theta)} F
    &= &\nabla_{q(\mathbf{s};\theta)} \ln \sum_s p(\mathbf{x,s};\theta) \\
                    &= &\nabla_{q(\mathbf{s};\theta)} \ln  \frac{\sum_s e^{-E(\mathbf{x,s;\theta})}}{\sum_{x,s} e^{-E(\mathbf{x,s\theta})}} \\
                    &= & \nabla_{q(\mathbf{s};\theta)} \Big( \underbrace{\ln \sum_s e^{-E(\mathbf{x,s;\theta})}}_{\textrm{utility }U} - \underbrace{\ln \sum_{x,s} e^{-E(\mathbf{x,s;\theta})}}_{\textrm{entropy }\mathcal{H}} \Big)
\end{eqnarray}

To find \(q(\mathbf{s})\), we apply a stochastic gradient descent on the negative free energy, and each iterative update step is as follows:

\begin{equation}
    \theta_{t} \leftarrow \theta_{t-1} - \frac{1}{A_\tau} \eta \sum_{A_\tau}  \nabla_{q(\mathbf{s};\theta)} -\mathcal{F}_{A_\tau} \hspace{2em} \forall A_\tau \in \mathcal{D}, \tau=1,...T
\end{equation}

where $\eta$ is the learning rate and the descent step of \(-\mathcal{F}\) represents the convergence towards a locally optimal variational approximation: \(q(\mathbf{s})=\prod_j q(s_j)\).
Depending on the form of the distribution (we used binary logistic distribution, i.e. \(q(s_j)=(1+e^{-Wx-c})^{-1}\) in our example), the optimization can be solved analytically.
Since the derivative can be inferred as the average energy change over \(A_\tau\), the gradient yields the difference between the expected utility $\ln \sum_s e^{-E(\mathbf{x,s};\theta)} = U$ and the entropy $\ln \sum_{x,s} e^{-E(\mathbf{x,s}\theta)} = \mathcal{H}$ gradients.

The utility $\ln \sum_s e^{-E(\mathbf{x,s};\theta)}$ is expressed as the energy \cref{eq:expected_energy} over all possible configurations of \(s\).
We can associate the first and second term as the expected energy value obtained from the conditional $p(\mathbf{s|x})$ and joint distribution $p(\mathbf{x,s})$ respectively (\ref{appx:algorithm}), using the gradient $\mathbf{W}$ as an example:

\begin{equation}
    \begin{aligned}
    \nabla_{q(\mathbf{s};\mathbf{W})} U &= \langle \mathbf{xs}\rangle_{p(\mathbf{s|x})} \\
    \nabla_{q(\mathbf{s};\mathbf{W})} \mathcal{H} &= \langle \mathbf{x s}\rangle_{p(\mathbf{x,s})}
    \end{aligned}
\end{equation}

The contrastive divergence (CD) algorithm takes a point estimate from one or more Gibbs sampling steps drawn to approximate the equilibrium energy:

\begin{equation}
    \begin{aligned}
    \langle \mathbf{x s}\rangle_{p(\mathbf{s|x})} &\sim \langle\mathbf{x}^0 \mathbf{s}^0\rangle \\
    \langle \mathbf{x s}\rangle_{p(\mathbf{x,s})} &\sim \langle\mathbf{x}^t \mathbf{s}^t\rangle
    \end{aligned}
\end{equation}

where $\langle\mathbf{x}^t \mathbf{s}^t\rangle$ is the average over product of the generated input samples multiplied and the generated latent variable samples from the Gibbs chain and \(\langle\mathbf{x}^0 \mathbf{s}^0\rangle\) is the initial sample (see pseudocode in \ref{appx:algorithm}).
Typically, a 1-step Gibbs sample chain $(CD_N; N=1)$ is sufficient for fast learning gradient estimation \citep{carreira2005contrastive}.
The gradient estimators can be used to minimize the objective function using a suitable learning rate.
The free energy is representative of the relative fit of the generative model with respect to the data distribution.
If the gap between the utility and entropy increases, it represents model overfitting \citep{hinton2012practical}.
A recent paper has highlighted a method of combining CD with variational inference which replaces the standard KL divergence, which shares similarities with RBMs and establishes the link between our method of estimation and standard variational inference \citep{ruiz2019contrastive}.

\section{Experiments}
\label{sec:experiments}
In this section, we describe the generative modelling process focusing on the data generation and inferring from the estimated latent variable component.
We describe how we pre-process the data and how the learning algorithm is used to optimize and generate statistically similar synthetic data for comparison.

\subsection{Case study}
\begin{figure}[!t]
    \centering
    \includegraphics[width=\textwidth]{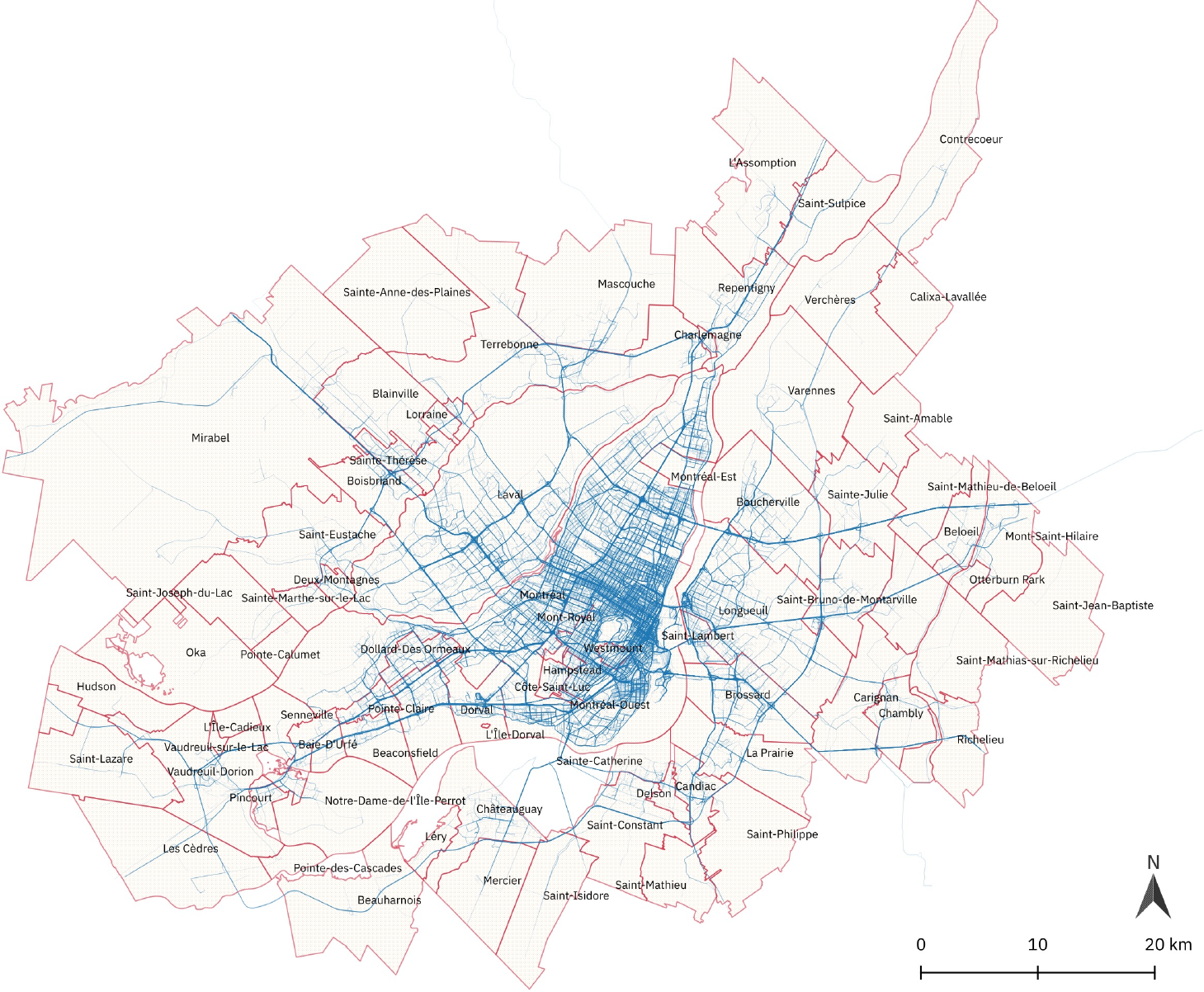}
    \caption{Visualization of the trip trajectories across the Greater Montr\'eal Region from the pilot study.}
    \label{fig:map1}
\end{figure}

We evaluate our proposed methodology on a trip trajectory dataset: the MTL Trajet GPS data from the Greater Montr\'eal Region \citep{datamobile2016}.
The open dataset consists of a total of 293,330 trip observations.
The data were collected from respondents living in the Greater Montr\'eal region (\cref{fig:map1}).
Trip trajectories were recorded in an application that runs in the background of participants' smartphone.
Participants were also prompted to report their travel mode and trip characteristics in addition to the GPS trajectories.
We consider the following revealed characteristics for our model: mode choice, trip purpose, trip distance, origin-destination point and departure/arrival time.

\subsection{Data pre-processing}
The GPS data from the mobile app are sampled at 4 to 10 second intervals.
From multiple users' GPS trajectories, we detect points at the origin and destination and matched to one of 34 boroughs of Greater Montr\'eal.
First, we verified each observation $D_n$ contains valid trajectory points, and we removed all corrupted data points outside the city boundary.
Next, we calculated the total trip distance between the start point and end point by the total sum of all point-to-point raw GPS coordinates.
Alternatively, open map data can also be used to find map matched travel distances.
Travel time was calculated by taking the time differential between the first and last coordinates.
Input time data $x_t$ were reparameterized into linear cyclic encoding features using sin/cosine transform: $x_{t_{sin}} = \sin(2\pi x_t)$, $x_{t_{cos}} = \cos(2\pi x_t)$.
Cyclic encoding features allow time data to be represented consistently and can be used as linear input.
Continuous data (trip distance, trip time) were normalized to unit variance.
Discrete categorical data (mode choice, purpose, origin-destination) were encoded as one-of-k vector: $x_{mode} = 2 \in \mathbb{R}^4 \rightarrow x_{mode_v} = \{0, 0, 1, 0\}$.
We selected trip candidates with a simple constraint of minimum 10-minute travel time and users had reported their travel mode and trip purpose.
Once all the valid trip observations were selected, we used this processed dataset for training and validation.
Since our methodology is an unsupervised learning algorithm, we did not consider any output data for cross-validation.
For model validation and data generation, we used the full training dataset to compare our results.

\subsection{Training}
We used a standard batch stochastic gradient descent learning algorithm for model estimation implemented using Theano Python machine learning libraries \citep{theano2016}.
The model parameters were updated after every batch sample.
We bootstrap iterations over mini-batches of observations, randomly sampled from the input data $\mathbf{x}_D$.
We defined a decaying learning rate $\eta$, starting at $10e^{-2}$ at the first iteration and decay at a rate of $0.1\%$ per batch.
The objective function is calculated as the difference in the first-order derivative of expected free energy of the input and the sampled data.
In this paper, we did not explore other novelty regularization methods such as dropout or model ensemble, which could be a future work for implementation.

\subsection{Data validation}
A typical estimation procedure would be to divide the data into training and validation sets.
The full dataset consists of a labelled subset (N=58,034) and an unlabelled subset (N=235,296).
The labelled subset consists of trips with full information availability and the unlabelled subset consist of trips with missing variables.
Using the labelled subset, we divide training and validation in a 70:30 ratio for model benchmarking against a comparable feedforward neural network (NN).
Accuracy validation is often misleading when a model is tested on a biased or imbalanced dataset.
In our case study, the dataset we obtained cannot fully represent the total population of the area due to physical limitations, e.g. availability of all transport modes, the use of smartphone applications, etc.
Therefore, we address this shortcoming by implementing a likelihood validation as a proxy to determine the model predictive accuracy.

For evaluating generative model performance, we simulate the model on the unlabelled data (with missing data) and compare the statistical properties of the generated output against the labelled dataset.
This is equivalent to testing the `unsupervised' learning performance.
The accuracy of these predictive probability distributions depends on whether the `correct' priors lead to reasonable predictive accuracy.
We estimated a series of models with different latent variable sizes and reported the model fit.
Ideally, increasing the size of latent variables would improve the fit for each variable dimension if input variables are assumed to be independent and identically distributed.
Our proposed method of variational Bayesian inference satisfies the likelihood principle where the inference depends on the distribution of the data \citep{berger1985}.

Next, we analyze the mean and variance effects of latent variables on the generative model.
Deep learning NN models are prone to overfitting when model parameters have a large bias and low variance, which result in poor predictors beyond the training data.
Such networks are naturally viewed as black-box functions and difficult to analyze.
By contrast, variational Bayesian inference allows the analyst to infer how flexible a model is warranted by the data \citep{mackay1995probable}.
Likewise, when parameters have low bias and high variance, it will result in low statistical confidence and makes the model harder to fit the data.
The consequence of the parameter uncertainty is that we cannot differentiate between good predictors and sampling error in our model.
Well-calibrated models should have flexibility in accounting for sampling error as well as robustness to avoid misspecification.

We also performed analysis over the elasticity of the choice probabilities w.r.t. to changes in the independent variables.
In our result, we show the direct elasticity of \textit{mode choice} with respect to \textit{travel distance}.
which can be calculated directly from the optimization step using the Jacobian function.

\subsection{Benchmarking}
We benchmark our results against a comparable single hidden layer feedforward NN with the number number of latent variables and mode choice as the output.
This is equivalent to partitioning the generative model into a hidden layer \(h(x)\) and computing the conditional output of the mode choice probability \(f(h(x))\).
The NN hidden and output layer equations are given by the following:
\begin{align}
    h(x) &= (1+e^{-(-\mathbf{xW}-\mathbf{c})})^{-1} = \sigma(-\mathbf{xW}-\mathbf{c})\\
    f(h(x)) &= \frac{e^{W_k h(x)+b_k}}{\sum_{k'} e^{W_{k'} h(x)+b_{k'}}}
\end{align}

The first difference between this approach and a discriminative-generative modelling approach (\ref{appx:generation}) is the direct estimation of the likelihood given the inputs, rather than an auxiliary step in generating latent variable samples than using these samples to generate the output mode choice data.
The second difference in the feedforward NN model is that the individual's observed utility is drawn from a non-linear deterministic component.
In contrast, the observed utility in the generative model is drawn from a linearly separable entropy term as described in \ref{appx:generation}.

\begin{figure}[!t]
    \centering
    \includegraphics[width=0.7\textwidth]{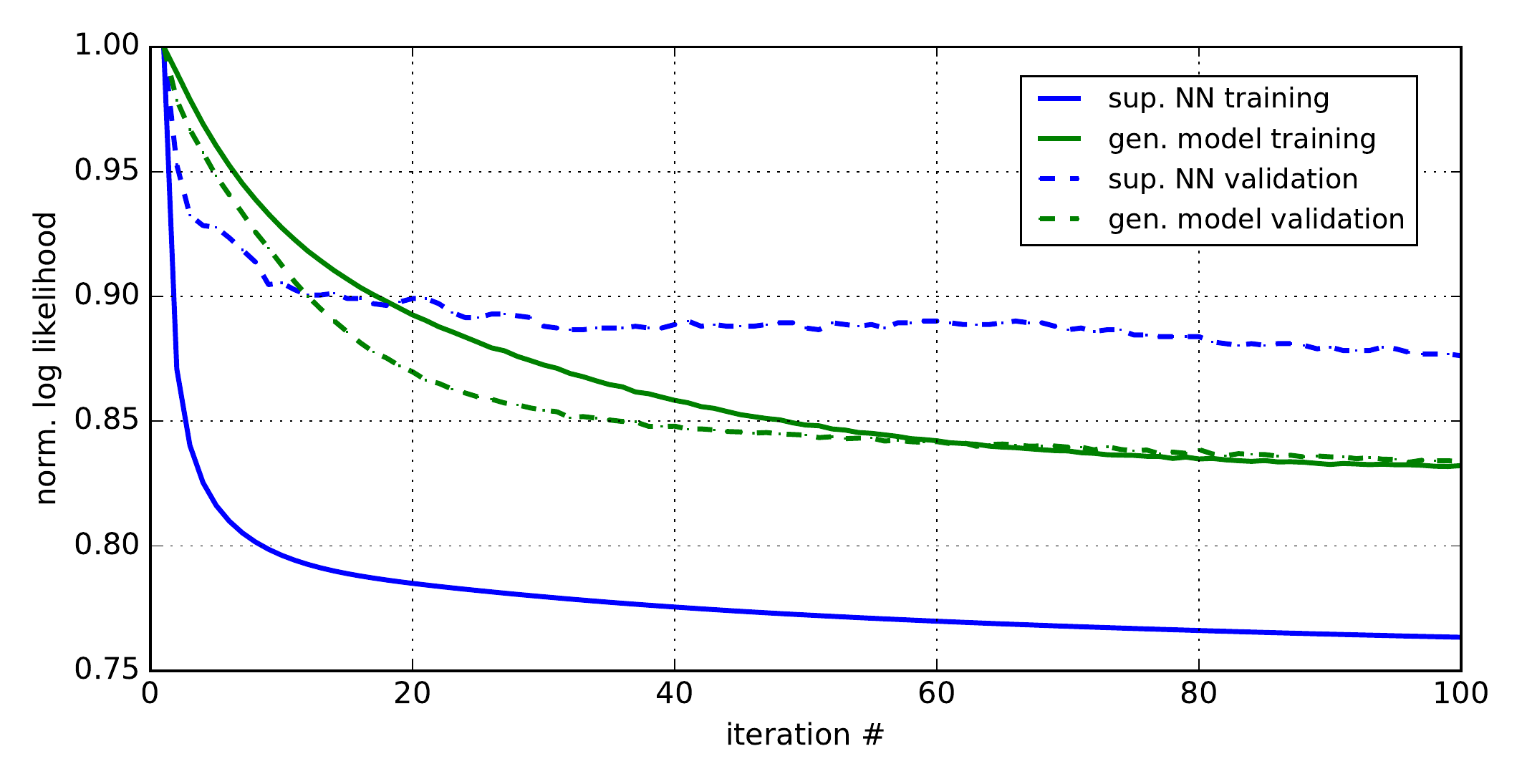}
    \caption{Training and validation likelihood curve (H=5)}
    \label{fig:training_curve_h5}
\end{figure}

\begin{figure}[!ht]
    \centering
    \includegraphics[width=0.7\textwidth]{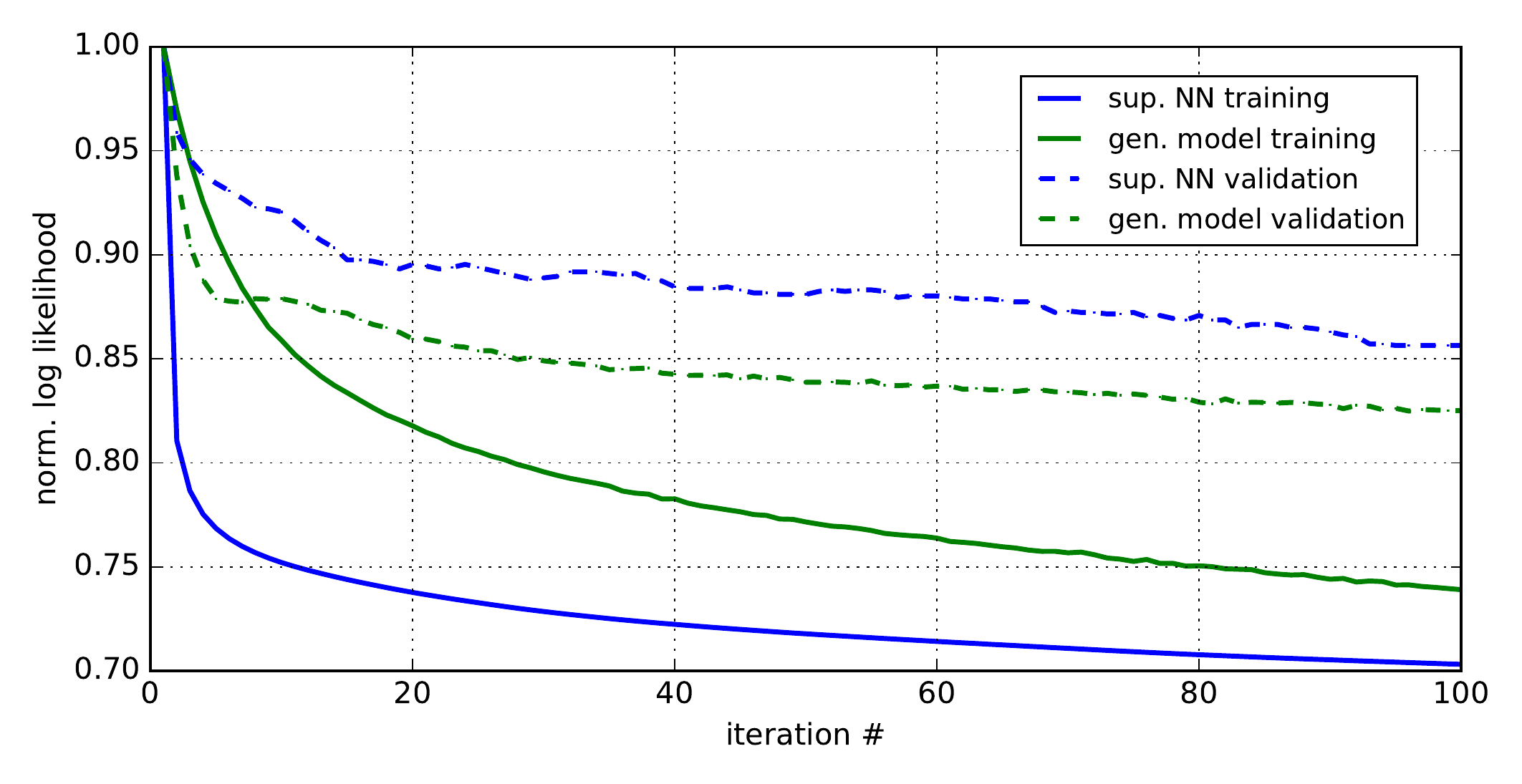}
    \caption{Training and validation likelihood curve (H=25)}
    \label{fig:training_curve_h25}
\end{figure}

\begin{figure}[!ht]
    \centering
    \includegraphics[width=0.7\textwidth]{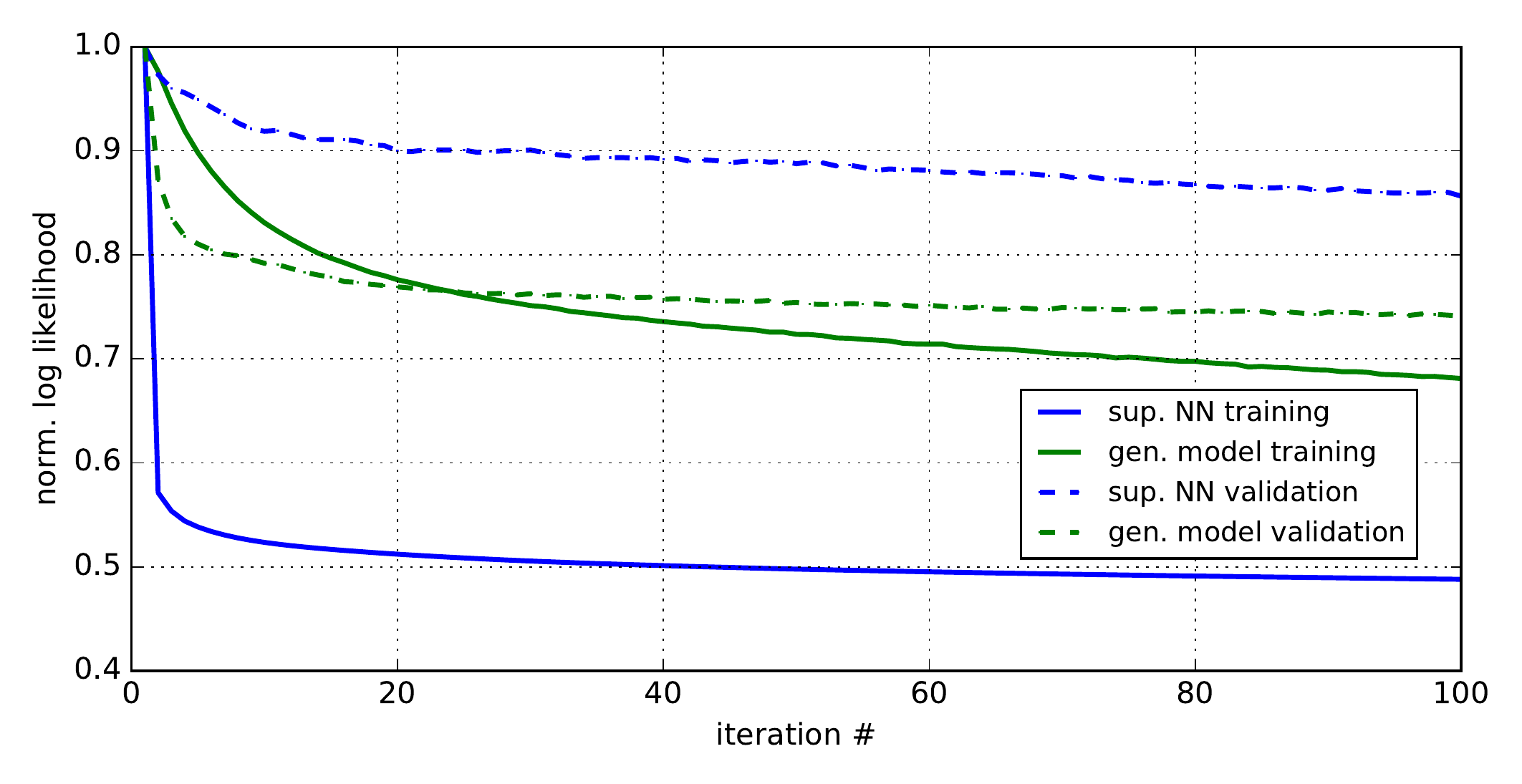}
    \caption{Training and validation likelihood curve (H=100)}
    \label{fig:training_curve_h100}
\end{figure}

We benchmark our model against the NN and compared the normalized log likelihood shown in \cref{fig:training_curve_h5,fig:training_curve_h25,fig:training_curve_h100}.
As expected, the \textit{training} curves converge asymptotically, which indicates that the gradient estimation reached a local optimum.
The \textit{validation} curves show the model fit on the validation data subset.
While the supervised NN training curve shows better model fit than the generative model in all 3 model instances (which is normal as the supervised NN model optimizes the model likelihood), it also points to higher overfitting shown by the more significant disparity between the training and validation likelihood.
Even though the generative model produces a weaker model fit on the training curve, the validation curve is better than the supervised NN and less likely to be overfitting.

\subsection{Latent constructs parameter analysis}
\begin{figure}[!t]
    \centering
    \includegraphics[trim=1cm 0 0.4cm 0, width=0.86\textwidth]{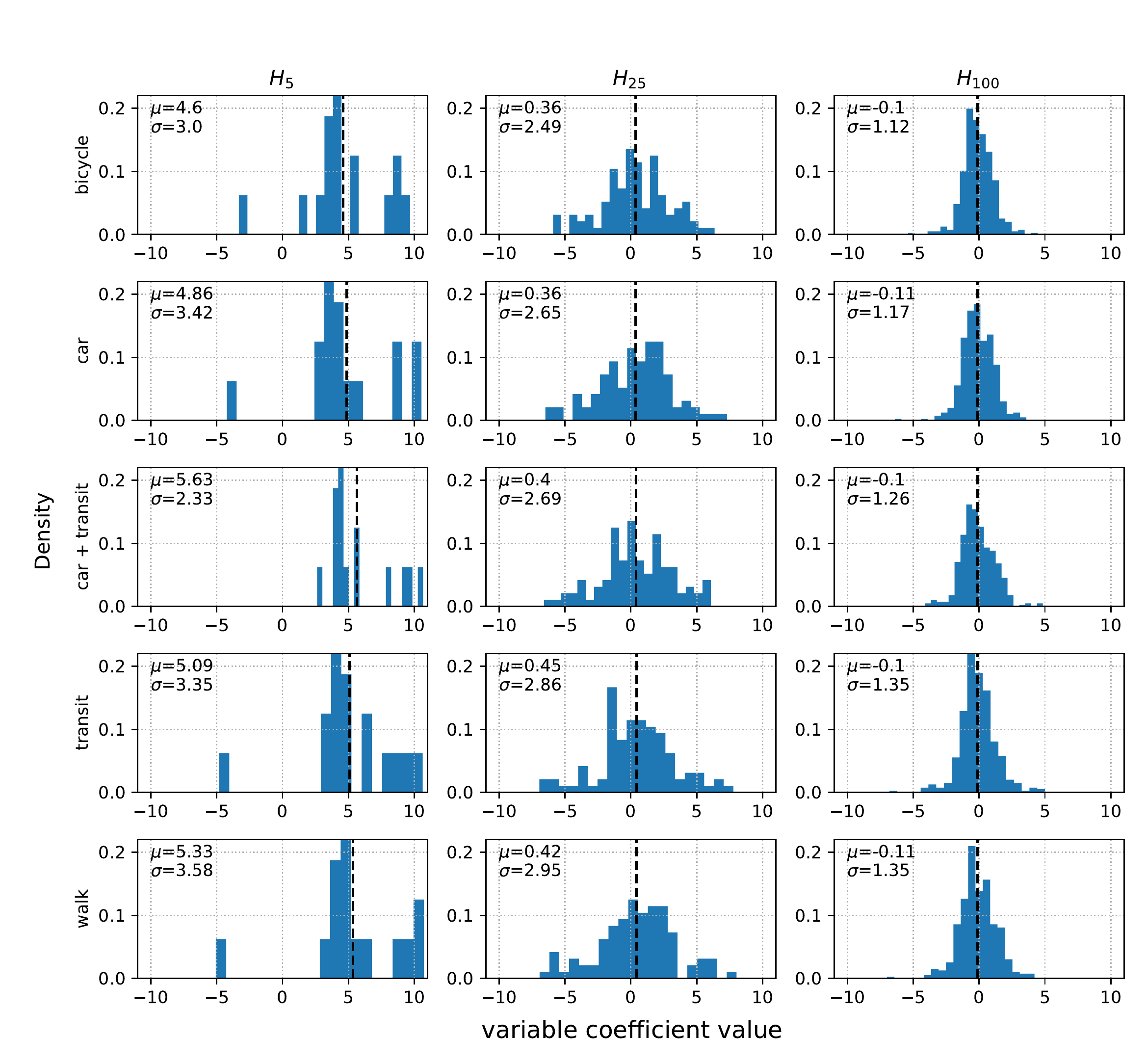}
    \caption{Histogram of parameter value distribution by mode choice and number of latent variables. Vertical dashed line represent distribution mean.}
    \label{fig:activation}
\end{figure}

For model analysis, we trained the model on a single layer fully connected network with $H=$ 5, 25 and 100 latent variables for 100 iterations over the dataset using our generative learning algorithm.
To verify if generative modelling provides better model generalization, we plot the distribution of the model parameters connecting the latent variables and mode choice data and computed the magnitude of mean and variance of the weight matrix.
The results are shown in \cref{fig:activation}.
We observed that with 5 latent variables $H_5$, the model parameters do not fit well to the input data.
The mean and variance parameter values are $H_{5}=\mathcal{N}(5.237, 9.33)$.
Increasing the number of latent variables substantially improves the model, where the mean and variance converges to zero mean and unit variance at $H_{25}$ and increasing to $H_{100}$ improve the model further.
The estimated mean and variance are $H_{25}=\mathcal{N}(0.45, 7.603)$ and $H_{100}=\mathcal{N}(-0.102, 1.624)$.

One reason for the improvement is the concept of sparse overcomplete representation of weights and activations in deep learning.
It has been shown that sparsity can be an important factor in explaining and capturing the variations in the data by reducing the number of activated parameters \citep{glorot11deep}.
The parameter distribution indicates the mean activation and utilization rate of latent variables.
When estimated parameters have low mean and variance, we can determine which subset of latent variables are `activated' and which are `inhibited' -- when parameters are zero or near zero, their contributions in the log domain is negative across the distribution.
This result suggests that the latent variables provide a strong indication of model identifiability by producing sparse parameter representation.


\subsection{Interpretation of latent constructs}
In conventional choice modelling latent variable interpretation are justified by explicitly introducing indicator variables to correspond to different latent variable states \citep{ashok2002extending}.
For example, a useful indicator might designate attitudinal variables: safety, comfortability or eco-friendliness \citep{vij2016and}.
However, for this method of latent variable classification to be useful, the indicators must be free from outliers and assumed to be uncorrelated to other events or error terms.
In generative modelling, we can emulate the travel decision process as a learning algorithm, to provide an underlying explanation for sensory information inputs. 
We can think of the latent variables as an interpretation of the observed data (e.g. how individuals consider their distance, mode choice, location choice, etc., simultaneously).

The earlier models that did not explicitly use psychometric indicators to capture the latent variables were alternative specific only and did not vary over the individual market segments \cite{ben2002integration}.
Our proposed model has no restrictions on latent variables being alternative specific.
It imposes a generalized logical structure (probabilistic graphical model) and accounts for uncertainty and variance from observed data (explanatory variables and observed choices) through Bayesian probability theory.
However, it is flexible enough that it can also be formulated as a model structure that only captures alternative specific variations by removing the connections between the latent constructs and the explanatory variables and any other setup is also possible.
The results indicated that while supervised learning performed better on the training set, it performed worse than the generative model on validation.
We modify the estimation step so that latent variables are conditioned on the connection strength between the observed and latent variables.
This representation can be more useful when attitudinal variables are not IID and have a high correlation with each other and thus, require knowledge of the underlying distribution.
The latent variable parameters define an entropy term which can be interpreted as a structure for capturing unobserved correlations between variables.
This can be framed as an entropy generalization to the linear MNL model structure where the latent variables form an error correction function.
This structure also represents a simplistic model of how decisions are simulated not just by random utility, but also the dynamical effects of information availability, habits and perceptions.
This reflects the role and importance of neural networks in capturing realistic behavioural responses beyond direct cause-and-effect maximum utility-based observations.

\subsection{Model elasticity}
\begin{figure}[!t]
  \centering
  \includegraphics[width=\textwidth]{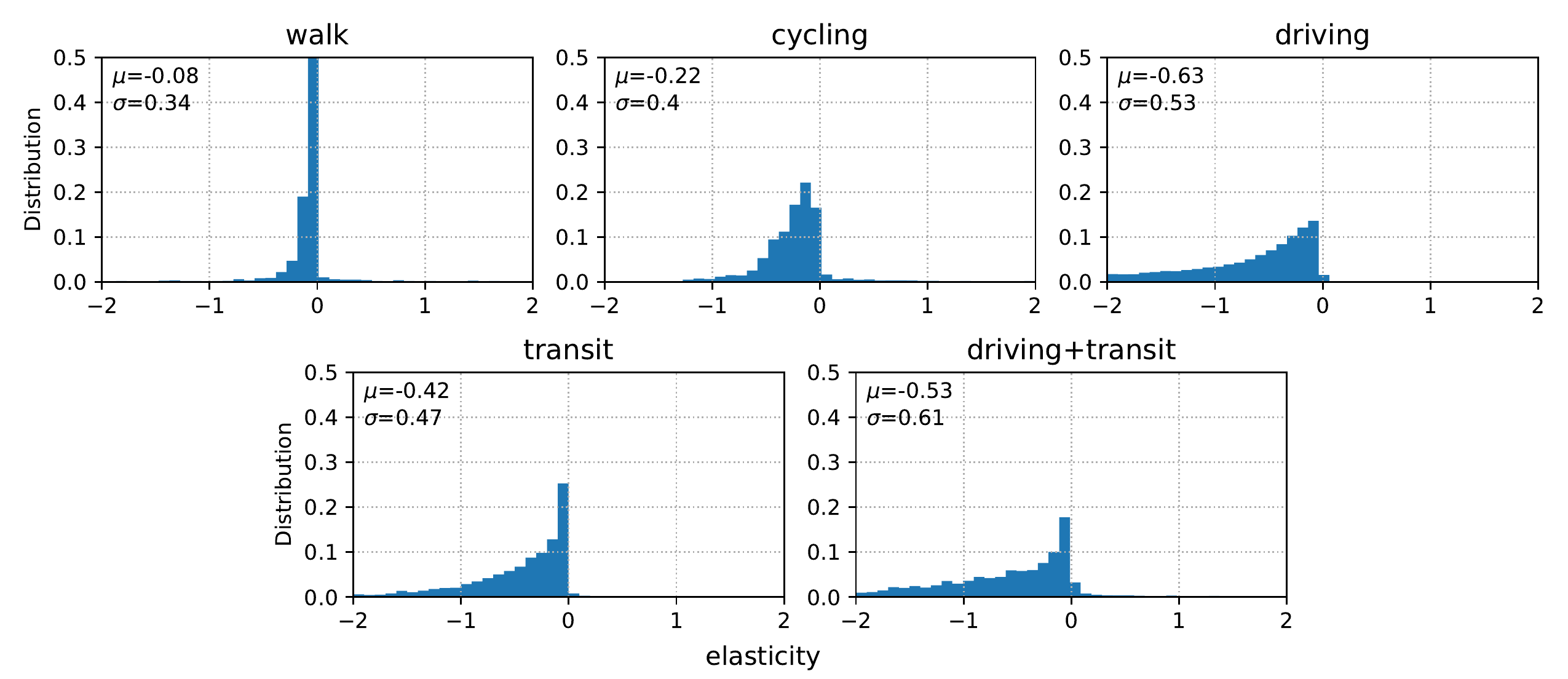}
  \caption{Distance elasticities on mode choice}
  \label{fig:elasticity}
\end{figure}

In econometric analysis, elasticity is an important metric to measure the effects of changes in the value of the explanatory variables (e.g. cost, distance) on choice probabilities. 
This test is an indicator of the variation in elasticities of the unobserved heterogeneity of the population w.r.t to the choice decision.
In the context of generative models, we can use the Jacobian determinant to compute the elasticity  (\ref{appx:elasticity}).
The direct elasticities of \textit{mode choice} with respect to \textit{travel distance} are shown in \cref{fig:elasticity}.
As expected, the elasticities are all negative. Distance is most strongly correlated with driving with an average elasticity of -0.635 and a standard deviation of 0.535.
Since walking trips are for relatively short distances, our results show that walking mode choice is inelastic w.r.t. distance with an average and standard deviation of -0.084 and 0.336 respectively.
Moreover, the average elasticity for driving mode is larger than transit or driving+transit mode, meaning that as the distance increases, the probability of driving decreases faster. This is verified by the collected GPS data, where individuals used public transit (commuter trains) more for long-distance trips -- especially for commuting.

Elasticity and latent variable parameter inspection can be regarded as measures of posterior and prior heterogeneity, respectively.
It puts forward a plausible model that assumes the generative model emulates an individual's prior information about the choice with respect to the latent variables, and the elasticity of demand for each mode choice in this context quantifies how much the individual would react to the decision having formed some prior beliefs generated from the model.

\subsection{Data simulation}
Generative models can be used to represent the underlying distribution of the data; thus, they can be instrumental in forecasting.
We used the trained model to generate samples from $p(\mathbf{x})$, which have similar statistical properties as the input data.
First, we consider a single observation where we observed only part of the data vector and the other part of the vector is unknown.
The unknown vector can be a single or multi-variable vector.
We denote this as $\mathbf{x}_D=(x_1,..., x_{D-1}, x_{D_{unk}})$, where $x_{D_{unk}}$ is fixed as the unknown variable.
The objective is to predict $x_{D_{unk}}$ using the remainder of the `known' data vector by sampling from the distribution $p(x_{D_{unk}}|x_1,..., x_{D-1})$.
We should note that conventional likelihood tests are not suitable in this instance because the outputs of the generative model are stochastic data-driven probability distributions, rather than a deterministic probability distribution of a dependent variable. \ref{appx:generation} describes how these distributions can be computed.

Next, we clamp the known variables to the input data and then sample the states of the hidden layer.
We use the sampled states of the hidden layer to generate the remaining state of the unknown variable, completing a full Gibbs sampling step.
This process is not limited to a single unknown variable.
If more unknown variables are used, it reduces the ability of the model to capture the data representation (this is analogous to adding noise to the input, we can fix $x_{D_{unk}}=0$ for the variables we want to forecast).
Therefore, the robustness of the model can be quantified by the information loss when adding noise to the input and how well it recovers this lost information.

\begin{figure}[!t]
    \centering
    \includegraphics[width=0.9\textwidth]{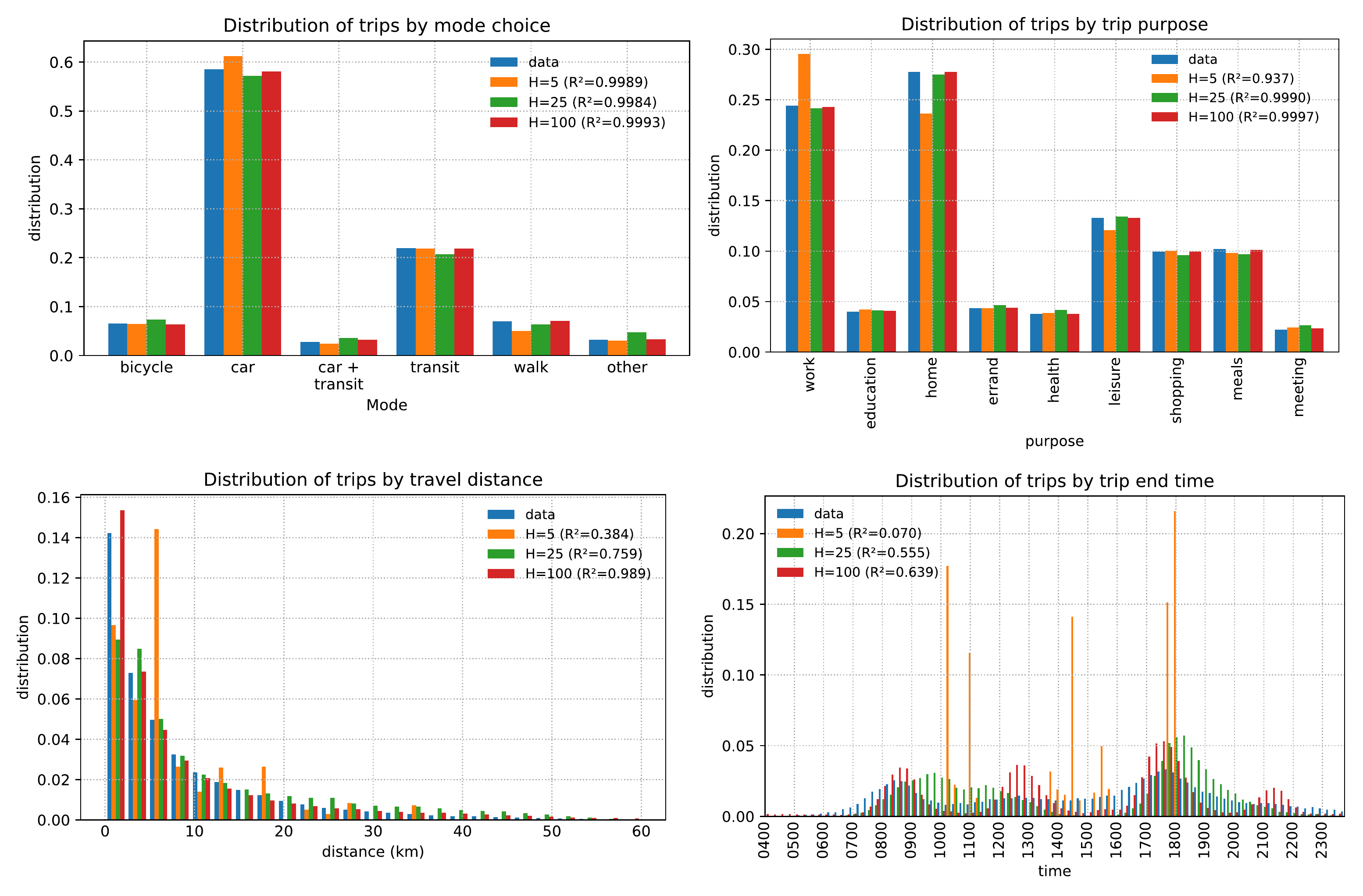}
    \caption{Discrete and continuous data generated from the model.}
    \label{fig:drawing}
\end{figure}

We show that as we increase the model capacity and complexity, the model can generate synthetic samples that emulate the original distribution of the data.
The output generated samples are evaluated against the inputs, and we compute the $R^2$ distribution fit.
The results are shown in \autoref{fig:drawing}.
In particular, we observed that discrete categorical variables (trip mode and trip purpose) are easily represented with small model capacity $(R^{2}>0.937)$, but continuous variables, e.g. trip distance $(H_{25}: R^2=0.759)$ and time $(H_{100}: R^2=0.639)$ require more latent variables to capture the underlying distribution accurately.

\begin{figure}[t]
    \centering
    \includegraphics[width=0.6\textwidth]{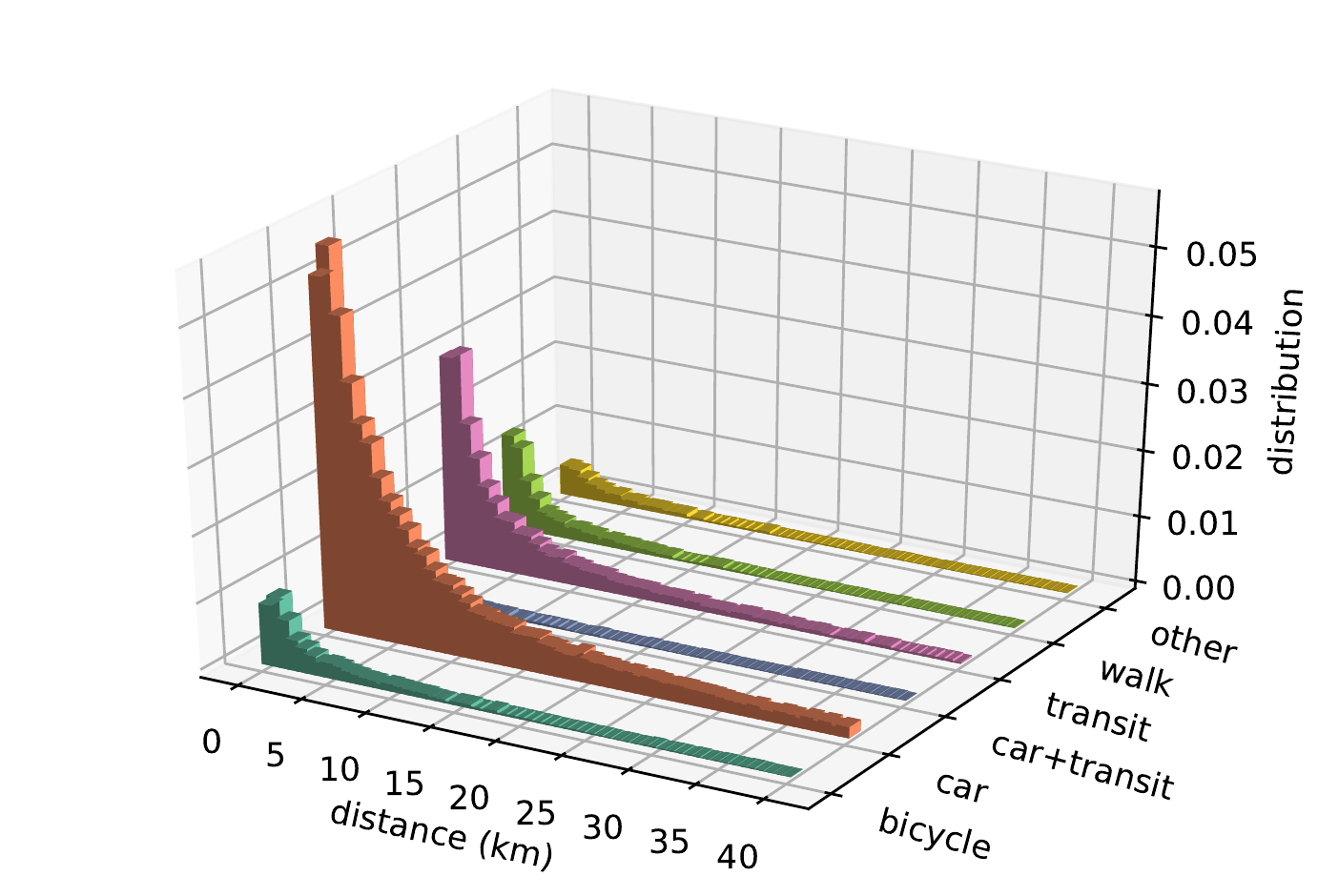}
    \caption{Data generation for a joint MDC output}
    \label{fig:cross_mode_dist}
\end{figure}

The generative model is also able to learn the multi-modal cyclical nature of trip arrivals, which is significantly challenging for a standard logit model to estimate.
In our simulation, the latent variables can generate a statistical distribution with modes in the morning and evening peak hours as well as a smaller peak around mid-afternoon.
Surprisingly, even with no indication of how the distribution is supposed to be or using any pre-defined measurement indicators, the generative model can capture the underlying properties of a complex distribution, demonstrating a level of understanding of the semantic variations in the dataset.
Finally, MD data can Cbe generated from the conditional probability densities -- an example would be combining mode choice with distance, as shown in \autoref{fig:cross_mode_dist}.

We analyze the model results using the Kruskal-Wallis statistical test and report the k-th central moments up to k=4 shown in \cref{tab:moment1,tab:moment2,tab:moment3,tab:moment4}. 
The results indicate that the samples generated from \(H=100\) are similar to the original data based on the test statistics and the p-values.
The k-th moments of the generated data converge to the k-th moments of the original data indicating that the generative model is well representative of the underlying behaviour.
We also report the variable pair correlation shown in \cref{tab:var_corr}. 
The correlation pair also confirms that the higher-order generative models can emulate the distribution of the original data with high accuracy.

The sample statistics of the generative model are shown in \cref{tab:stat_test}.
For each of these models, we report the two-way likelihood Chi-square test, mean squared distance and p-value of the generative models on discrete variables mode and trip purpose.
For trip distance and trip arrival counts, we report the RMSE of the samples against the original data.
RMSE for trip distance are \((H_{5}:4.171, H_{25}:4.721 \textrm{ and } H_{100}:1.852)\).
RMSE for trip arrival counts are \((H_{5}:128.7, H_{25}:26.8 \textrm{ and } H_{100}:23.5)\)
Model significance is computed as the p-values for \(\chi^2\) at 5\% sample size.
The analysis shows that the models with a larger number of latent variables are more consistent and statistically significant (\(H=100: \chi^2=2.3308, p\leq 0.115\)),  even though \(R^2\) values indicate that the generative model well represents the data.

As shown in the results, the generative model can represent both discrete and continuous data types simultaneously.
This relates to the sparsity concept mentioned in the previous subsection -- the model is robust to corrupted data and information retrieval from truncated data are possible.
This experiment shows how we can use a generative model for model prediction and forecasting for various input variable types.
In terms of latent variables, this is not an exhaustive analysis, and we can increase the size of latent variables to increase the representational power, but with diminishing returns.
However, it has been shown that for a neural network with $T$ input dimensions and $T-1$ latent variables, it is globally stable and satisfy the necessary conditions for optimality with no local minima in the error surface \citep{yu1992can}.

While these tests may serve as useful benchmarks, we note that the choice of latent variable size is still arbitrary and dependent of many various factors including data size, number of variables, complexity and amount of `missing' information in the data collection.
However, as we have shown that in general, generative modelling may serve as a useful additional tool for travel behaviour analysts to estimate MDC data using variational Bayesian inference techniques.
Collectively our analysis of the generative modelling provides empirical support that unobserved information in the data plays an important role in the model estimation, which has previously shown to be plausible in discrete choice theory \citep{matejka2015rational}.

\section{Conclusion}
As the use of machine learning models and algorithms become increasingly significant and essential in travel behaviour research, more emphasis has to be put on model interpretability rather than pure forecast accuracy.
Our work focuses on methods and tools for analyzing and interpreting complex travel behaviour data and estimation of MDC models.
Notably, we introduced a generative machine learning approach for analyzing and estimating large scale MDC travel behaviour model that uses variational Bayesian inference for model training.
We proposed an RBM-based learning algorithm to model behaviour data, accounting for information heterogeneity and variable correlations.
Our main contribution relates to model specification and model identification of MDC data using our RBM based algorithm that uses variational inference.
We showed how the proposed model could be used to compute the conditional probability distribution of the dependent variables as well as the associated elasticities.
This concept can be expressed in terms of information gain to quantify their contribution to utilitarian behaviour by measuring the KL divergence between the observed and simulated data.

For the case study, we implemented the algorithm on an open large travel behaviour dataset.
We were able to estimate model parameters to fit the underlying distribution of the data while retaining identifiability and sparsity.
The sparse distribution of parameters enabled the generative model to capture the correlation effects between input variables for both discrete and continuous variable types.
To ensure that latent variables capture data heterogeneity, simulation tests were performed, and we showed that the generative model was able to recover the original data with similar statistical distribution.
For model interpretability, we show that elasticities can be obtained for economic analysis.
Also, we report model statistics, correlation and sample analysis, which indicate that the shape of the distribution converges to the original data samples.

We note that the additional complexity due to an increase in the hidden units is minimal when using the first-order stochastic gradient optimization.
The increased size of model parameters did not constitute a significant increase in computational time, due to fast and efficient tensor-based operations using Theano machine learning libraries.
We also observed that increasing the number of hidden units of an order of magnitude does not correspond to the same increase in computational time.
On the other hand, increasing the number of observations in the training data will be the main bottleneck in each iteration of the dataset.
In future work, we can also look at regularization, e.g. L1/L2 penalty or drop out techniques, to reduce overfitting from the effects of using latent variables.

With the development of ubiquitous data collection methods for travel behaviour analysis, there are potentials for generative machine learning to be used for modelling these large multidimensional travel information datasets.
Overall, integrating probabilistic variational Bayesian inference methods can improve model tractability and interpretability.
Adopting this framework into dynamic road pricing, route choice recommendation and traffic network simulation are some interesting applications for future work.

\section*{Acknowledgement}
This research is funded by the Canada Research Chair in Disruptive Transportation Technologies and Services and Ryerson University.

\clearpage
\appendix
\section{Multiple discrete and continuous conditional probability generation}
\label{appx:generation}
We provide additional details here on how conditional probability generations are formulated.
To apply a generative model to travel behaviour choice problems, we have to specify the distribution of our required output variable set conditioned on the other variables.
We can also further extend this to other distributions, not just multinomial and Gaussians, e.g. unimodal distribution for ordinal data \citep{da2008unimodal}.

\begin{figure}[!b]
  \centering
  \includegraphics{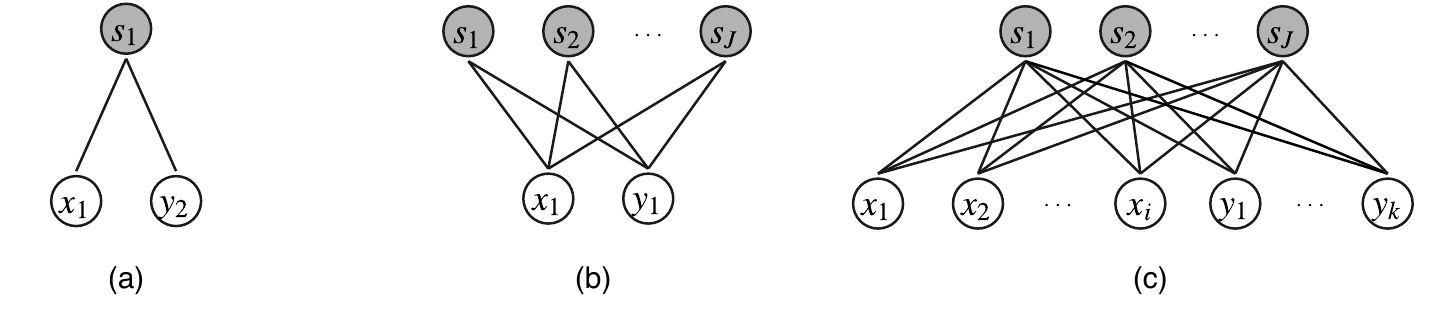}
  \caption{Generating various multiple discrete-continuous outputs using generative models. (a) Example 1, (b) Example 2, (c) Example 3.}
  \label{fig:appx_mdc}
\end{figure}

\begin{exm}
  We first consider the simplest possible example consisting of two observed variables $ [x_1, y_1]$ connected by a single hidden unit $s_j$ (\hyperref[fig:appx_mdc]{Figure \ref{fig:appx_mdc}a}).
  The generative model captures the joint distribution of $x, y$ and $s$ expressed as $P(x,y,s) = \frac{1}{Z}e^{-E(x,y,s)}$ as derived from \autoref{eq:joint_pdf}.
  The functional form that represents the variables under an RBM energy model is $E(x,y,s)=-\sum_{s_j}x_1W_{1,j}s_j -\sum_{s_j}y_1W_{1,j}s_j -b_1x_1 -\sum_{s_j}c_js_j -d_2y_1 $ and the conditional probability of $y$ given $x$ assuming $y$ is a multinomial output:

  \[
    P(y_1|x_1) = \frac{e^{-F(x_1,y_1)}}{\sum_{y_1'}e^{-F(x_1,y_1')}}
  \]

  where its variational free energy $F(x_1,y_1)$ is calculated as:

  \begin{eqnarray*}
    F(x_1,y_1) = -\ln \sum_{s_j\in\{0,1\}} e^{-E(x_1,y_1,s_j)} &=& -b_1x_1 -d_2y_1 -\ln\sum_{s_j\in\{0,1\}}e^{-s(x_1W_{1,j}+y_1W_{1,j}+c_j)} \\
    &=& -b_1x_1 -d_2y_1 -\ln(1+e^{-x_1W_{1,j}-y_1W_{1,j}-c_j})
  \end{eqnarray*}

\end{exm}

The first term $b_1x_1$ is the `error-corrected' utility component in the model.
However, unlike in conventional DCM, $b_1$ is the beta of variable $x_1$ contribution to the full \textit{joint} probability $P(x,y,s)$.
The second term can be interpreted as the `alternative specific constant' (ASC) of $y_1$.
For instance, if $y_1$ is a 3-alternative discrete variable $y_1:\{y_1^1, y_1^2, y_1^3\}$, then $d_2$ is a 3-dimension vector representing the ASCs.
In the conditional probability $P(y_1|x_1)$, if $y_1^1 =1$ and $0$ otherwise, then the error-corrected utility of alternative $y_1^1$ is:

\begin{eqnarray*}
  F(x_1,y_1^1) &=& -\Big(b_1x_1 + d_2^1 \cdot (y_1^1 =1)
                   + d_2^2 \cdot (y_1^2 =0) + d_2^3 \cdot (y_1^3 =0)
                   + \ln(1+e^{ - x_1W_{1,j} - y_1W_{1,j} - c_j})\Big) \\
               &=& -\Big(b_1x_1 + d_2^1 + \underbrace{\ln(1+e^{- x_1W_{1,j}
                   - y_1W_{1,j} - c_j})}_{\textrm{single correction term}}
                   \Big)
\end{eqnarray*}

If the weights connections to the hidden units are reduced to zero, i.e. $W_1=0, W_2=0$ and $c_j=0$, then the model collapses into a standard MNL.
For such a configuration:

\[
  F(x_1,y_1^1) = -\Big(b_1x_1 + d_2^1 + \ln(1 + e^{0})\Big)
               = -\underbrace{(b_1x_1 + d_2^1)}_{\textrm{MNL utility}}
\]

\begin{exm}
  We consider the same example above, but expanding to $j$ hidden units $s_1,...,s_j$. With $j$ hidden units, additive terms are added to the error-corrected utility (\hyperref[fig:appx_mdc]{Figure \ref{fig:appx_mdc}b}):

  \[
    F(x_1,y_1^1) = -\Big(b_1x_1 + d_2^1 + \underbrace{\sum_{j}\ln(1
                   + e^{- x_1W_{1,j} - y_1W_{1,j} - c_j})}_{\textrm{
                   multiple correction terms}}\Big)
  \]

\end{exm}

\begin{exm}
  Lastly, we consider multiple inputs and multiple discrete-continuous outputs: The joint probability expands to $i$ input variables $x_1,...x_i$ (\hyperref[fig:appx_mdc]{Figure \ref{fig:appx_mdc}c}).
  Likewise, the error-corrected utility can be derived as:

  \[
    F(x_1,...x_i,y_1,...,y_k) = -\Big(\sum_i b_ix_i + d_2^1
                          +\sum_{j}\ln(1+e^{-\sum_ix_iW_{i,j}-\sum_ky_kW_{k,j}-c_j})
                          \Big)
  \]

\end{exm}

\noindent where $y_k$ can be any discrete or continuous variable.
These examples above can be extended to multiple discrete-continuous joint distributions, where each $y_k$ component is a Product of Experts model:

\[
    P(y_1,...y_k|x_1,...x_i) = \prod_k P(y_k|x_1,...x_i)
\]

\noindent We also note here that the correction terms are \textit{marginal decreasing} functions for $x_i \rightarrow \infty$ and $W_{i,j} > 0$,

\[
  \lim_{{x_i,...x_i}\to\infty} F(x_1,...x_i,y_1,...,y_k)=-(\sum_i b_ix_i + d_2^1)\
  \hspace{0.5em}\implies W_{i,j} > 0
\]

\noindent For continuous variable output with positive only values, the stepped sigmoidal function is applied to $F(x_1,...x_i,y_{\textrm{cont}})$:

\begin{equation*}
  f(y_{\textrm{cont}}|x_1,...x_i) = \ln(1+e^{-F(x_1,...x_i,y_1,...,y_k)})
\end{equation*}

\noindent If the output is linear with range $-\infty < y_1 < \infty$, then the output would be the variational free energy:

\begin{equation*}
  f(y_{\textrm{linear}}|x_1,...x_i) = F(x_1,...x_i,y_1,...,y_k)
      = \sum_i \frac{(b_i - x_i)^2}{2} - d_2 -
        \sum_{j}\ln(1+e^{-\sum_ix_iW_{i,j}-\sum_ky_kW_{k,j}-c_j})
\end{equation*}

\noindent For discrete choice outputs, a similar method described in Example 1 and 2 is used:
\[
    P(y_{\textrm{discrete}}|x_1,...x_i) = \frac{e^{-F(x_1,...x_,y_{\textrm{discrete}})}}{\sum_{y_{\textrm{discrete}}'}e^{-F(x_1,...x_,y_{\textrm{discrete}}')}}
\]

\pagebreak
\section{RBM learning algorithm}
\label{appx:algorithm}
  \begin{algorithm}[!h]
    \SetAlgoLined
    \caption{RBM learning algorithm for generative modelling using N-step Gibbs sample chain $(CD_N)$}
    \label{alg:rbm}
    \SetKwInOut{Input}{Input}
    \SetKwInOut{Output}{Output}

    \Input{
      RBM data sample $\mathcal{D} = \{\mathbf{x}_1,...,\mathbf{x}_n\}$,
      batch sample $A_i\subset\mathcal{D},i=1,...,d$,
      learning rate $\eta$,
      iteration steps $T$}
    \Output{gradient approximation $\theta=(\mathbf{W},\mathbf{c},\mathbf{b}).$}
    \BlankLine
    init: $\theta=0$, $\tau=1$\;
    \ForAll{$A_\tau\in\mathcal{D}, \tau=1,...,T$}
    {
      \ForAll{$(\mathbf{x}_n)\in A_\tau$}
      {
        \For{$t=1$ \KwTo $N$}
        {
          $CD_t$: iterate over Gibbs chain\\
          \BlankLine
          positive phase\\
          $\mathbf{x}^0\leftarrow\mathbf{x}_n$\\
          $\mathbf{s}^0\sim \prod_{j=1}^H p(s_j|\mathbf{x}^0)$\\
          \BlankLine
          negative phase\\
          $\mathbf{x}^t\sim \prod_{i=1}^I p(x_i|\mathbf{s}^0)$\\
          $\mathbf{s}^t\sim \prod_{j=1}^H p(s_j|\mathbf{x}^t)$\\
        }
      }
      \% Variational free energy term \\
      $\nabla_{q(\mathbf{s};\theta)} (-\mathcal{F})_{A_\tau} \approx (\langle\mathbf{x}^t\mathbf{s}^t\rangle - \langle\mathbf{x}^0\mathbf{s}^0\rangle$) \\
      \% parameter update step\\
      \For{$\theta\in\theta$}
      {
        $\theta_{\tau+1}\leftarrow\theta_{\tau}
        -\eta\nabla_{q(\mathbf{s};\theta)} (-\mathcal{F})_{A_\tau}$\;
      }
    }
  \end{algorithm}
  
\section{Case study experiment}
\label{appx:case_study}
For the case study example described in \cref{sec:experiments}, we show the derivations of the joint probability, energy function and estimation steps:

\subsection*{Energy function}

The model is defined by the following energy function:

\begin{equation}
    E(\mathbf{x,s},y) = -\Big(b^y y + \sum_{m,j} x_m W_{m,j} s_j + \sum_{j} y W_{j} s_j + \sum_m b^x_m x_m+ \sum_j c_j s_j \Big)
\end{equation}

where \(s_j \in \{0, 1\}^J\) and \(x_m,y \in \mathbb{R}^{\mathcal{D}} \) are referred to as latent and observed variables respectively in the RBM model. 
\(m\) is the number of explanatory variables used, \(x_m\) are the explanatory variables (time, speed, distance, location etc.) and \(y\) is the mode choice dependent variable vector.
For 5 latent variables, we set \(j=5\).
The weight parameters are \(\theta=(W, b^x, b^y, c)\).

\subsection*{Joint probability}
The joint probability distribution of the observed and hidden variables follows the Boltzmann distribution \(p(\mathbf{x,s},y) =e^{-E(\mathbf{x,s},y)}/Z\), where Z is a normalization factor such that \(0 < p(\mathbf{x,s},y) \leq 1\).

\subsection*{Model estimation process}
Given a sufficient number of latent variables, we can tune the RBM model parameters such that we minimize the negative free energy:

\begin{equation}
    \begin{aligned}
    &(-F) &= &\ln \sum_{s_j\in\{0,1\}}\Big(e^{b^y y + \sum_{m,j} x_m W_{m,j} s_j + \sum_{j} y W_{j} s_j + \sum_m b^x_m x_m+ \sum_j c_j s_j}\Big) - \ln Z
    \end{aligned}
\end{equation}

The training task is then to minimize the negative free energy term by taking the derivative w.r.t. the model parameters and updating the parameters using an SGD training process.
The gradient update step is as follows:
\begin{enumerate}
    \item Draw Gibbs samples \(\mathbf{x}^0, \mathbf{y}^0, \mathbf{s}^0, ...,\mathbf{x}^t, \mathbf{y}^t, \mathbf{s}^t\) for \(t\) steps (\ref{appx:algorithm}).
    \item Compute \(\frac{\partial}{\partial \theta} \Big(-F(\mathbf{x}^0, \mathbf{y}^0, \mathbf{s}^0, ...,\mathbf{x}^t, \mathbf{y}^t, \mathbf{s}^t)\) \Big)and update model parameters \(\theta\).
\end{enumerate}

The model can be used to predict new observations by ``clamping'' the explanatory variables, generate latent variable samples from it, and using the generated samples to compute the choice probability \(P(y|\mathbf{x,s})\).
  
\section{Model elasticity}
\label{appx:elasticity}
Analyzing model elasticity is a way to test functional dependency among a set of observations $n$ on the conditional probability distribution.
For these tests we exploit the computational graph used to calculate the backpropagation algorithm in stochastic gradient descent by substituting the final partial derivative ${\partial \hat h}/{\partial \mathbf{W}}$ with ${\partial \hat h}/{\partial x_n}$.
The advantage of using a Jacobian is that it allows discrimination of linear and non-linear dependence in the model.
We generate the Jacobian matrix for each example of the conditional output on the set of inputs and estimate the density of elasticities across the data points.

\begin{lem}
  Given the conditional probability function $p_{n}(\mathbf{x})$, its elasticity $\varepsilon$ is defined as:

    \[
      \varepsilon = \frac{Jp_{n}(\mathbf{x}) \mathbf{x}_n}{p_{n}(\mathbf{x})}
                  = \frac{\partial p_{n}(\mathbf{x})}{\partial x_n} \cdot
                    \frac{\mathbf{x}_n}{p_{n}(\mathbf{x})}
    \]

  The Jacobian matrix $Jp_{n}(\mathbf{x})$ of each observation $n$, for each fixed input vector $\mathbf{x}$ is defined as the backpropagation derivative w.r.t $p_{n}$:

  \begin{eqnarray*}
    Let\hspace{1em} p_n(\mathbf{x}) &=& g(\mathbf{W}^{(1)} \cdot h(\mathbf{W}^{(0)}\cdot \mathbf{x}_n)), \hspace{1em}then\\
    Jp_{n}(\mathbf{x}) &=& \frac{\partial p_{n}(\mathbf{x})}{\partial x_n} = \underbrace{\frac{\partial p_{n}(\mathbf{x})}{\partial \hat h}\cdot \frac{\partial \hat h}{\partial x_n}}_{\textrm{backpropagation terms}} = \begin{bmatrix}
    \frac{\partial p(\mathbf{x})_1}{\partial \hat{h}_1} & \dots & \frac{\partial p(\mathbf{x})_1}{\partial \hat{h}_s} \\
    \vdots & \ddots & \vdots\\
    \frac{\partial p(\mathbf{x})_k}{\partial \hat{h}_1} & \dots & \frac{\partial p(\mathbf{x})_k}{\partial \hat{h}_s}
    \end{bmatrix} \cdot
    \begin{bmatrix}
    \frac{\partial \hat{h}_1}{\partial x_1} & \dots & \frac{\partial \hat{h}_1}{\partial x_n} \\
    \vdots & \ddots & \vdots\\
    \frac{\partial \hat{h}_s}{\partial x_1} & \dots & \frac{\hat{h}_s}{\partial x_n}
    \end{bmatrix}
  \end{eqnarray*}

\end{lem}

\clearpage
\section{Model Analysis Results}
\cref{tab:moment1,tab:moment2,tab:moment3,tab:moment4} reports the k-th central moments of the original and generated data up to k=4 and the Kruskal-Wallis statistical tests.
We report the variable correlations in the original and generated data in \cref{tab:var_corr}.
The correlation analysis shows high similarities between the original data and model \(H=100\).
Sample statistics from the data generated by the models are shown in \cref{tab:stat_test}.

\begin{table}[h]
    \centering
    \caption{k-th central moment of the original data samples}
    \begin{tabu}{X[0.9] X[1,R] X[1,R] X[1,R] X[1,R]}
        \toprule
        Moment & mode & purpose & distance & time \\
        \midrule
        1 & 0 & 0 & 0 & 0\\
        2 & 1.56 & 6.28  & 109.62   & 2.518e+05\\
        3 & 1.78 & 5.32   & 2961.60   & 9.848e+07 \\
        4 & 6.80 & 70.23 & 1.62e+05 & 1.923e+11 \\
        \midrule
        Kruskal-Wallis & - & - & - & - \\
        p-value  & - & - & - & - \\
        \bottomrule
    \end{tabu}
    \label{tab:moment1}
\end{table}

\begin{table}[h]
    \centering
    \caption{k-th central moment of the generated data samples (H=5)}
    \begin{tabu}{X[0.9] X[1,R] X[1,R] X[1,R] X[1,R]}
        \toprule
        Moment & mode & purpose & distance & time \\
        \midrule
        1 & 0 & 0 & 0 & 0 \\
        2 & 1.23 & 6.21 & 47.35 & 4.931e+06 \\
        3 & 0.95 & 12.13 & 642.88 & 4.497e+10 \\
        4 & 2.70 & 86.86 & 1.598e+04 & 4.702e+14 \\
        \midrule
        Kruskal-Wallis & 513.94 & 1165.31 & 510.98 & 85.13 \\
        p-value  & \(\leq 0.05\) & \(\leq 0.05\) & \(\leq 0.05\) & \(\leq 0.05\)\\
        \bottomrule
    \end{tabu}
    \label{tab:moment2}
\end{table}

\begin{table}[h]
    \centering
    \caption{k-th central moment of the generated data samples (H=25)}
    \begin{tabu}{X[0.9] X[1,R] X[1,R] X[1,R] X[1,R]}
        \toprule
        Moment & mode & purpose & distance & time \\
        \midrule
        1 & 0 & 0 & 0 & 0 \\
        2 & 1.62 & 6.23 & 152.98 & 7.058e+05 \\
        3 & 1.84 & 5.25 & 2765.74 & 9.100e+08 \\
        4 & 7.25 & 68.85 & 1.019e+05 & 2.435e+12 \\
        \midrule
        Kruskal-Wallis & 1.01 & 1.31 & 326.58 & 6.74 \\
        p-value  & \(\leq 0.315\) & \(\leq 0.253\) & \(\leq 0.05\) & \(\leq 0.05\)\\
        \bottomrule
    \end{tabu}
    \label{tab:moment3}
\end{table}

\pagebreak

\begin{table}[h]
    \centering
    \caption{k-th central moment of the generated data samples (H=100)}
    \begin{tabu}{X[0.9] X[1,R] X[1,R] X[1,R] X[1,R]}
        \toprule
        Moment & mode & purpose & distance & time \\
        \midrule
        1 & 0 & 0 & 0 & 0 \\
        2 & 1.58 & 6.28 & 131.94 & 5.983e+05 \\
        3 & 1.81 & 5.33 & 3280.01 & 7.743e+08 \\
        4 & 6.94 & 70.02 & 1.356e+05 & 1.767e+12 \\
        \midrule
        Kruskal-Wallis & 0.393 & 0.092 & 20.88 & 1.63 \\
        p-value  & \(\leq 0.53\) & \(\leq 0.76\) & \(\leq 0.05\) & \(\leq 0.2\)\\
        \bottomrule
    \end{tabu}
    \label{tab:moment4}
\end{table}

\begin{table}[h]
    \centering
    \caption{Variable pair correlation}
    \begin{tabu}{X[1.5] X[1,R] X[1,R] X[1,R] X[1,R]}
        \toprule
        Variable pair &  Original data & H=5 & H=25 & H=100 \\
        \midrule
        mode-purpose      & -0.0961 & -0.1149 & -0.1002 & -0.0954 \\
        mode-distance     & -0.1884 & -0.4439 & -0.2269 & -0.1919 \\
        mode-time         &  0.0349 & -0.0244 &  0.0667 &  0.0382 \\
        purpose-distance  & -0.1396 & -0.2846 & -0.1549 & -0.1453 \\
        purpose-time      & -0.1039 & -0.3866 & -0.1504 & -0.1052 \\
        distance-time     &  0.4777 &  0.8247 &  0.5715 &  0.4907 \\
        \midrule
        mean difference   &      -  &  0.07   & -0.004 & -0.001   \\
        \bottomrule
    \end{tabu}
    \label{tab:var_corr}
\end{table}

\begin{table}[h]
    \centering
    \caption{5\% sample size analysis of the generative choice model outputs}
    \begin{tabu}{ X[1.8] X[1,R] X[1,R] X[1,R] X[1,R] }
        \toprule
        Model & \(\chi^2\) & dist. & \(R^2\) & p-value \\
        \midrule
         & \multicolumn{4}{c}{mode choice} \\
        \cline{2-5} 
        H=5 & 23.658 & 4.6791 & 0.9989 & 1.0 \\
        H=25 & 33.8029 & 6.1074 & 0.9984 & 1.0 \\
        H=100 & 2.3308 & 1.5569 & 0.9993 & \(p\leq 0.115\) \\
        \cline{2-5} 
        \vspace{0.5em}
         & \multicolumn{4}{c}{trip purpose} \\ 
          \cline{2-5}
        H=5 & 55.535 & 7.5163 & 0.937 & 1.0 \\
        H=25 & 6.041 & 2.499 & 0.9994 & \(p\leq 0.36\) \\
        H=100 & 0.334 & 0.582 & 0.9997 & \(p\leq 0.01\) \\
        \bottomrule
    \end{tabu}
    \label{tab:stat_test}
\end{table}

\pagebreak

\bibliographystyle{elsarticle-harv}

\bibliography{bibliography}

\end{document}